\documentclass[11pt]{article}
\usepackage{coling2018}
\usepackage{times}
\usepackage{url}
\usepackage{latexsym}

\usepackage{url}
\usepackage{multirow}
\usepackage{wrapfig}
\usepackage{subfig}
\usepackage{graphicx}

\usepackage{enumitem}
\usepackage[draft,backgroundcolor=yellow,textsize=scriptsize]{todonotes}

\newcounter{tbsnr}   
\newenvironment{tbs}   
{\addtocounter{tbsnr}{1}\par\bigskip \noindent\fbox{\thetbsnr}   
\hspace*{\fill}\begin{minipage}{7cm}\tt}   
{\end{minipage}\hspace*{\fill}\bigskip}   
\newcommand{\tb}[1]{\begin{tbs}{#1}\end{tbs}}

\newcommand{\cut}[1]{}

\newcommand{\rf}[1]{\textcolor{blue}{RF: #1}}
\newcommand{\rb}[1]{\textcolor{blue}{\textbf{RB: #1}}}
\newcommand{\rs}[1]{\textcolor{red}{RS: #1}}
\definecolor{awesome}{rgb}{1.0, 0.13, 0.32}
\newcommand{\eb}[1]{\textcolor{awesome}{\textbf{EB: #1}}}
\newcommand{\baum}[1]{\textcolor{green}{\textbf{TB: #1}}}
\newcommand{\aash}[1]{\textcolor{orange}{\textbf{AASH: #1}}}

\title{\emph{Ask No More:} Deciding when to guess in referential visual dialogue}

\author{Ravi Shekhar$^\dagger$, Tim Baumg\"{a}rtner$^*$, Aashish Venkatesh$^*$,\\ 
\textbf{Elia Bruni$^*$,
  Raffaella Bernardi$^\dagger$} and \textbf{Raquel Fernandez$^*$}\\
$^*$University of Amsterdam, $^\dagger$University of Trento \\
 {\tt raquel.fernandez@uva.nl}  \ \ \ {\tt raffaella.bernardi@unitn.it}}

\date{}

\begin{document}
\maketitle

\begin{abstract}
Our goal is to explore how the abilities brought in by a dialogue manager can be included in end-to-end visually grounded conversational agents. 
We make initial steps towards this general goal by augmenting a task-oriented visual dialogue model with a decision-making component that decides whether to ask a follow-up question to identify a target referent in an image, or to stop the conversation to make a guess.  Our analyses show that adding a decision making component produces dialogues that are less repetitive and that include fewer unnecessary questions, thus potentially leading to more efficient and less unnatural interactions.
\end{abstract}
\blfootnote{First 4 authors contributed equally.}

%

\section{Introduction}

\blfootnote{This work is licensed under a Creative Commons Attribution 4.0 International License. License
details:\\ \url{http://creativecommons.org/licenses/by/4.0/}\\
Data and code are available at  \url{https://vista-unitn-uva.github.io}}


The field of interactive conversational agents, also called dialogue
systems, is receiving renewed attention not only within Computational
Linguistics (CL) and Natural Language Processing (NLP) -- its original
and probably most natural locus -- but also within the Machine
Learning (ML) and the Computer Vision (CV) communities. The
overarching challenge, in line with the long-term aims of Artificial
Intelligence, is to develop data-driven agents that are capable of
perceiving (and possibly acting upon) the external world and that we
can collaborate with through natural language dialogue to achieve
common goals.

\begin{wrapfigure}{R}{0.48\textwidth}
\begin{minipage}{3cm}
\includegraphics[width=2.7cm]{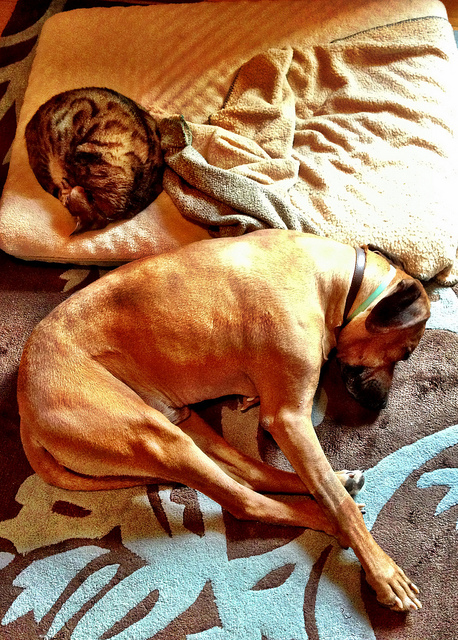} 
\end{minipage}
\begin{minipage}{3cm}\small
\begin{tabular}[t]{l@{\hspace{-0.2cm}}r}
\bf Questioner & \bf Answerer \\[3pt]
1. Is it a person? & No\\[3pt]
2. Is it the dog?  & Yes\\[3pt]
\multicolumn{2}{l}{\bf $\leadsto$ \it success by our model} \\[4pt]
3. The dog? & Yes\\[3pt]
4. Is it in the foreground? & No \\[3pt]
5. Is it the whole dog? & Yes\\[3pt] 
\multicolumn{2}{l}{\bf $\leadsto$ \it success by baseline model}
\end{tabular}
\end{minipage}\caption{Dialogue that leads to task success in the {\it GuessWhat?!}~game by our model, which decides when to stop asking questions, and by the baseline model in \newcite{guesswhat_game}, which does not.\label{fig:needofus}}
\vspace*{-.3cm}
\end{wrapfigure}

Within the ML and CV communities, recent research on conversational
agents combined with Deep Learning techniques has yielded interesting
results on visually grounded  tasks~\cite{visdial,guesswhat_game,imag:mosta17}. In this line of research, the focus is mostly on improving model performance by investigating new machine learning paradigms (like reinforcement learning
or adversarial learning) in end-to-end settings, where the model learns directly from raw data without symbolic annotations ~\cite{stru:end17,best:lu17,arey:wu17}.
Task accuracy, however, is not the only criterion by which a conversational agent should be judged. Crucially, the dialogue should be coherent, with no unnatural repetitions nor unnecessary questions --- unlike the 5-turn dialogue shown in Figure~\ref{fig:needofus}. To achieve this, a conversational agent needs to learn a strategy to decide how to respond given the current context and the task at hand. These abilities are typically considered part of \emph{dialogue management} and have been the focus of attention in dialogue systems research within the CL/NLP
community \cite{LarssonTraum2000,williams2008partially,bohus2009ravenclaw,young2013pomdp}.

In this paper, we thus take a step back: instead of focusing on learning paradigms, we focus on the system architecture.  
We argue that the time is ripe for exploring how the abilities brought in by a dialogue manager can be included in end-to-end 
conversational agents within the
Deep Learning paradigm mostly put forward by the ML and CV communities.
We make initial steps towards this general goal by augmenting the 
task-oriented visual dialogue model proposed by~\newcite{guesswhat_game}, not yet with a full-fledged dialogue manager, but with a decision-making component that decides whether to ask a follow-up question to identify a target referent in an image, or to stop the conversation to make a guess (see Figure~\ref{fig:needofus}).
Our focus is on providing a thorough analysis of the resulting dialogues. Our results show that 
 the presence of a decision making component leads to dialogues that are less repetitive and that include fewer unnecessary questions.

%

\section{Related Work}
\label{sec:related}

Our system operates on both linguistic and visual information. 
Visually-grounded dialogue has experienced a boost in recent years, in part thanks to the construction of large visual human-human dialogue datasets built by the Computer Vision  community \cite{imag:mosta17,visdial,guesswhat_game}.
These datasets include two participants, a Questioner and an Answerer, who ask and answer questions about an image. For example, in the \emph{GuessWhat?!}~dataset developed by \newcite{guesswhat_game}, which we exploit in the present work, a Questioner agent needs to guess a target object in a visual scene by asking yes-no questions (more details are provided in the next section).

Research on visually-grounded dialogue within the Computer Vision community exploits encoder-decoder architectures~\cite{sutskever2014sequence} --- which have shown some promise for modelling chatbot-style dialogue~\cite{VinyalsLe:2015,sordoni-EtAl:2015:NAACL-HLT,serban2016building,Li-EtAl:NAACL2016,Li-EtAl:emnlp2016} --- augmented with visual features. This community has mostly focused on model learning paradigms. 
Initial models, proposed by~\newcite{guesswhat_game} and \newcite{visdial}, use supervised
  learning (SL): the Questioner and the Answerer are trained to
generate utterances (by word sampling) that are similar to the human gold standard. 
To account for the intuition that dialogues require some form of planning, subsequent work by \newcite{visdial_rl} and \newcite{stru:end17}
makes use of reinforcement learning (RL). 
In all these approaches but~\newcite{stru:end17}, however, the Questioner performs a non-linguistic action (i.e., selects an image or object within an image) after a fixed number of question-answer rounds. Thus, there is no decision making on whether  further questions are or are not needed to identify a visual target. To address this limitation, \newcite{stru:end17} put forward a more flexible approach: They let the Questioner ask at most 8 questions, but introduce an extra token ({\tt stop})
within the vocabulary, which the question generation model has to learn.
This strategy, however, is suboptimal: The question generator needs to generate probabilities for items that
  do not lie on the same distribution (the distribution of natural
  language words vs.~the distribution of binary decisions
  \emph{ask}/\emph{guess}).\footnote{We also note that on the \emph{GuessWhat?!}~GitHub page at \url{https://github.com/GuessWhatGame/guesswhat} it is mentioned that in the updated version of the system by \newcite{stru:end17} ``qgen [the question generator] stop learning to stop'' (GitHub accessed on 16/03/2018).}
 Our work addresses this
limitation in a more principled way, by including a new decision-making module within the encoder-decoder architecture and analysing its impact on the resulting dialogues.

We build on work by the dialogue systems community. 
In traditional dialogue systems, the basic system architecture includes several
components -- mainly, a language interpreter, a dialogue
manager, and a response generator -- as discrete modules that operate
in a pipeline~\cite{jurafksy-martin-dialogue,jokinen2009spoken} or in
a cascading incremental
manner~\cite{SchlangenSkantze2009eacl,dethlefs2012optimising}. The
\emph{dialogue manager} is the core component of a dialogue agent: it
integrates the semantic content produced by the interpretation
module into the agent's representation of the context (the {\em
  dialogue state}) and determines the next action to be performed by
the agent, which is transformed into
linguistic output by the generation module. Conceptually, a dialogue manager thus includes
  both (i) a {\em dialogue state tracker}, which acts as a context model that ideally keeps track of aspects such as current goals, commitments made in the dialogue, entities mentioned, and the level of shared understanding among the participants \cite{clark1996using}; and (ii) an  {\em action selection policy}, which makes decisions on how to act next, given the current dialogue state. In the present work, we focus on incorporating a \emph{decision-making module} akin to an action selection policy into a visually-grounded encoder-decoder architecture and leave the integration of other more advanced dialogue management aspects for future work.

In particular, work on incremental dialogue processing, where a system needs to decide not only \emph{what} to respond but also \emph{when} to act \cite{rieser-schlangen2011}, has some similarities with the problem we address in the present paper, namely, when to stop asking questions to guess a target.\footnote{Our system is not word-by-word incremental at this point, but given the incremental nature of encoder-decoder architectures, an extension in this direction should be possible. We leave this for future work.} 
Researchers within the dialogue systems community have applied different approaches to design incremental dialogue policies for how and when to act. Two common approaches are the use of rules parametrised by thresholds that are optimised with human-human data \cite{buss2010collaborating,ghigi2014incremental,PaetzelEtal2015sigdial,kennington2016supporting} and the use of reinforcement learning \cite{kim2014inverse,khouzaimi2015optimising,manuvinakurike2017using}.  For example, \newcite{PaetzelEtal2015sigdial} implement an agent that aims to identify a target image out of a set of images given descriptive content by its dialogue partner. Decision making is handled by means of a parametrised rule-based policy: the agent keeps waiting for additional descriptive input until either her confidence on a possible referent exceeds a given threshold or a maximum-time threshold is reached (in which case the agent gives up). The thresholds are set up by optimising points per second on a corpus of human-human dialogues (pairs of participants score a point for each correct guess). In a follow-up paper by \newcite{manuvinakurike2017using}, the agent's policy is learned with reinforcement learning, achieving higher performance. 

 
We develop a decision-making module that determines, after each question-answer pair in the visually grounded dialogue, whether to ask a further question or to pick a referent in a visual scene. We are interested in investigating the impact of such a module in an architecture that can be trained end-to-end directly from raw data, without specific annotations commonly used in dialogue systems, such as dialogue acts \cite{PaetzelEtal2015sigdial,manuvinakurike2017using,kennington2016supporting}, segment labels \cite{manuvinakurike2016real}, dialogue state features \cite{williams2013state-tracking,young2013pomdp,kim2014inverse}, or logical formulas \cite{yu2016training}.

\section{Dataset}
\label{sec:guesswhat}

To develop our model and perform our analyses, we use the 
\emph{GuessWhat?!}~dataset,\footnote{Available at: \url{https://guesswhat.ai/}}
a dataset of approximately 155k human-human dialogues created via Amazon Mechanical
Turk~\cite{guesswhat_game}.  \emph{GuessWhat?!}~is a cooperative two-player
game: both players see an image with several objects; one player (the Oracle) is assigned
a target object in the image and the other player (the Questioner) has to
guess it. To do so, the Questioner has to ask Yes/No
questions to the Oracle. When the Questioner thinks he/she can guess the object, the list of
objects is provided and if the Questioner picks the right one the game is
considered successful. No time limit is given, but the
Questioner can leave the game incomplete (viz.~not try to guess).
The set of images and target objects has been built from the training and validation sections of the
MS-COCO dataset~\cite{lin:micr14} by only keeping
images that contain at least three and at most twenty
objects and by only considering target objects whose
area is big enough to be located well by humans ($area >500px^2$). 
Further details are provided in Appendix A.

\section{Models}
\label{sec:models}

\newcite{guesswhat_game} develop models of the Questioner and Oracle roles in \emph{GuessWhat?!}.  
We first describe their models, which we consider as our baseline, and then describe our modified Questioner model. 
As explained below, \newcite{guesswhat_game} model the Questioner role by means of two disconnected modules: a Question Generator (QGen) and a Guesser, that are trained independently. After a fixed number of questions by QGen, the Guesser selects a candidate object. We propose and evaluate a model of the Questioner role that incorporates a decision-making component that connects the tasks of asking and guessing (which we take to be part of the planning capabilities of a single agent) and offers more flexibility regarding the number of questions asked to solve the game.

%

\begin{figure}\centering
\begin{minipage}{9.8cm}\centering
\includegraphics[width=9.5cm]{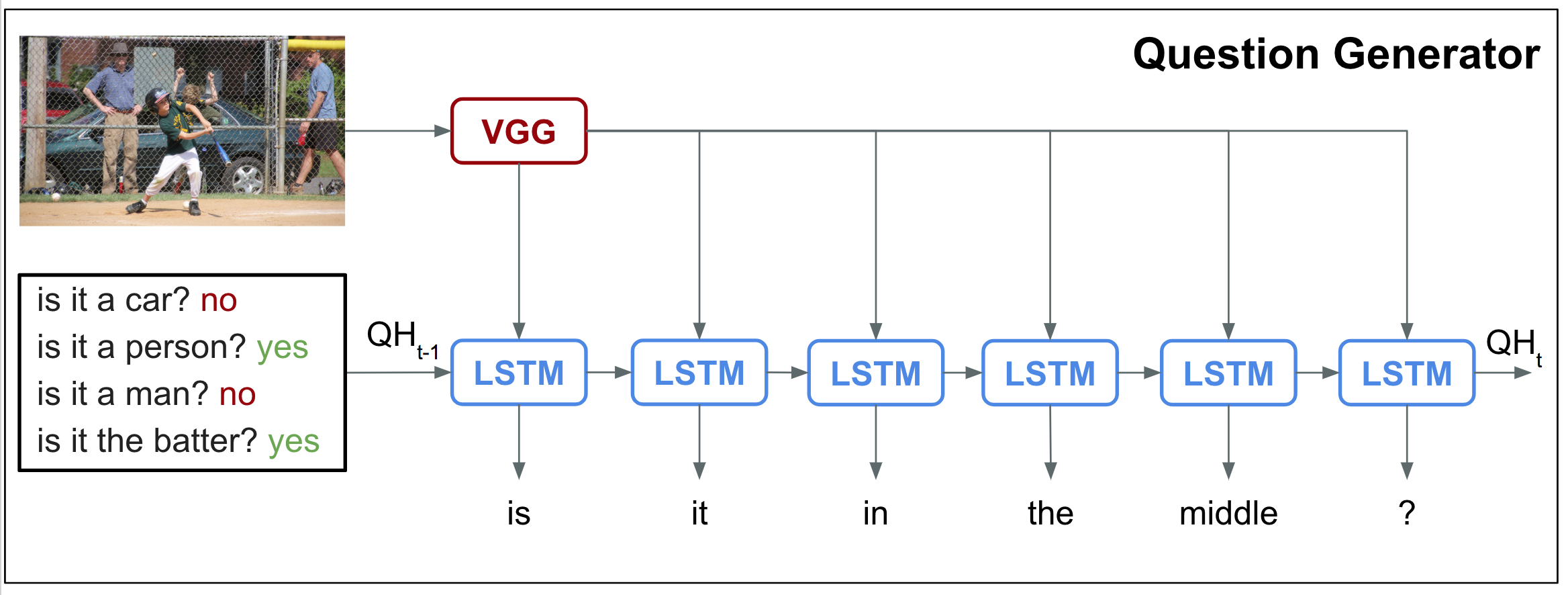}\\
\includegraphics[width=9.5cm]{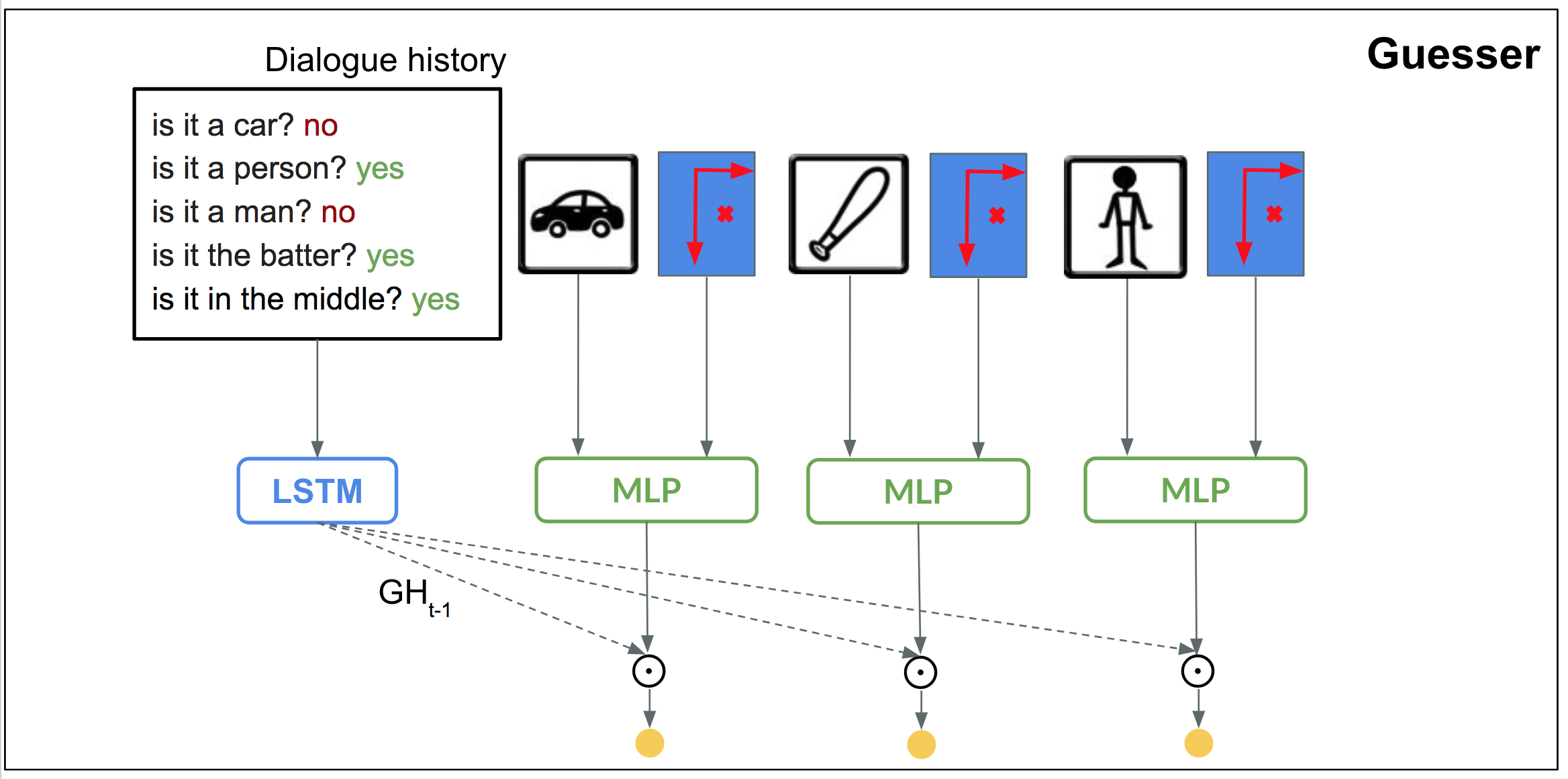}
\end{minipage}
\begin{minipage}{5.3cm}\centering
\includegraphics[width=5cm]{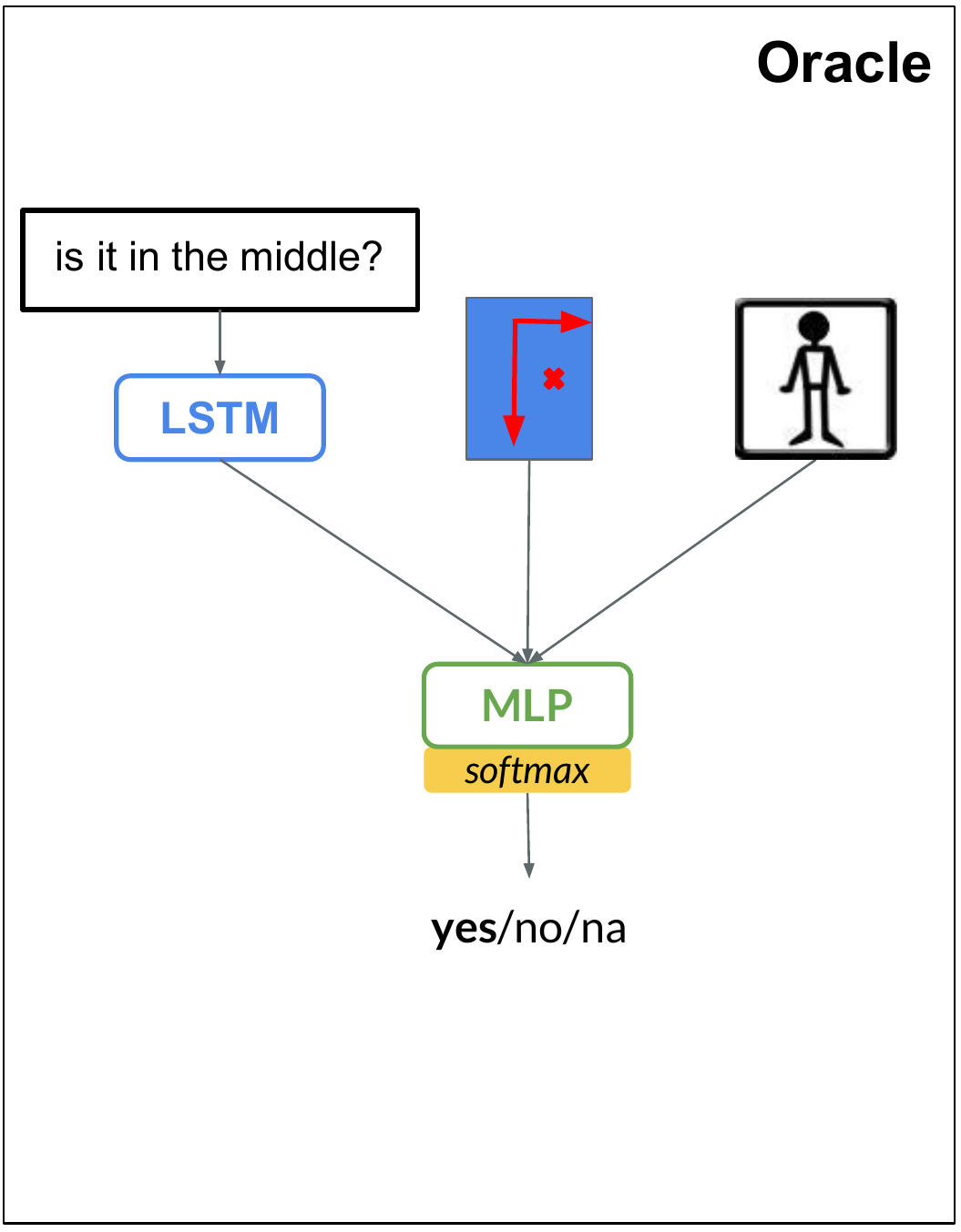}
\end{minipage}
\caption{Baseline models. Lefthand side: Independent modules for the Questioner model: Question Generator (top) and Guesser (bottom). Righthand side: Oracle model. 
\label{fig:qgen-guesser-oracle}}
\end{figure}

\subsection{Baseline Models}
\label{sec:baseline}

We provide a brief description of the models by \newcite{guesswhat_game}, which we re-implement for our study.
Further details on the implementation of each module are
available in Appendix A.

\paragraph{Question Generator (QGen)} This module is implemented as a Recurrent Neural Network
(RNN) with a transition function handled with Long-Short-Term Memory
(LSTM), on which a probabilistic sequence model is built with a Softmax
classifier. Given the overall image (encoded by extracting its VGG features) and the current dialogue history (i.e. the previous sequence of
questions and answers), QGen produces a
representation of the visually grounded dialogue (the RNN's hidden
state $QH_{t-1}$ at time $t-1$ in the dialogue) that encodes information useful to generate the next
question $q_t$. See the sketch on the top-right part of Figure~\ref{fig:qgen-guesser-oracle}.

\paragraph{Guesser} The best performing model of the Guesser module by \newcite{guesswhat_game}
represents candidate objects by their object category and 
spatial coordinates. These features are passed through a Multi-Layer Perceptron (MLP) to get an embedding for each object.
The Guesser also takes as input the dialogue history processed by an LSTM, whose hidden state $GH_{t-1}$ is of the same size as the MLP output.
A dot product between both returns a score for each candidate object in the image. A diagram of the architecture is given on the bottom-right section of Figure~\ref{fig:qgen-guesser-oracle}.


\paragraph{Oracle} The Oracle is aware of the target object and answers each question by QGen with Yes, No, or Not Applicable. 
The best performing model of the Oracle module by \newcite{guesswhat_game}
takes as input embeddings of the target object category, its spatial coordinates, and the current 
question. These embeddings are concatenated into a single vector and fed
to an MLP that outputs the answer, as illustrated in Figure~\ref{fig:qgen-guesser-oracle}, righthand side.

\subsection{Our Questioner Model}
\label{sec:our_model}

We extend the Questioner model of~\newcite{guesswhat_game} with a third
module, a decision making component (DM) that determines, after each question/answer pair, whether QGen should {\em ask} another question or whether the Guesser should {\em guess} the target object. We treat this
decision problem as a binary classification task, for which we use an
MLP followed by a Softmax function that outputs probabilities for the two classes of interest: \emph{ask} and \emph{guess}.
The Argmax function then determines the class of the next action. With this approach, we bypass the need to specify any decision thresholds and instead let the model learn whether enough evidence has been accumulated during the dialogue so far  to let the Guesser pick up a referent.\footnote{Note also that, since we have separate states for  \emph{ask} and \emph{guess} decisions, we could potentially extend the approach to decide among multiple action types beyond the binary case. }

\begin{wrapfigure}{R}{9.5cm}\centering
\vspace*{-5pt}
\includegraphics[width=9.5cm]{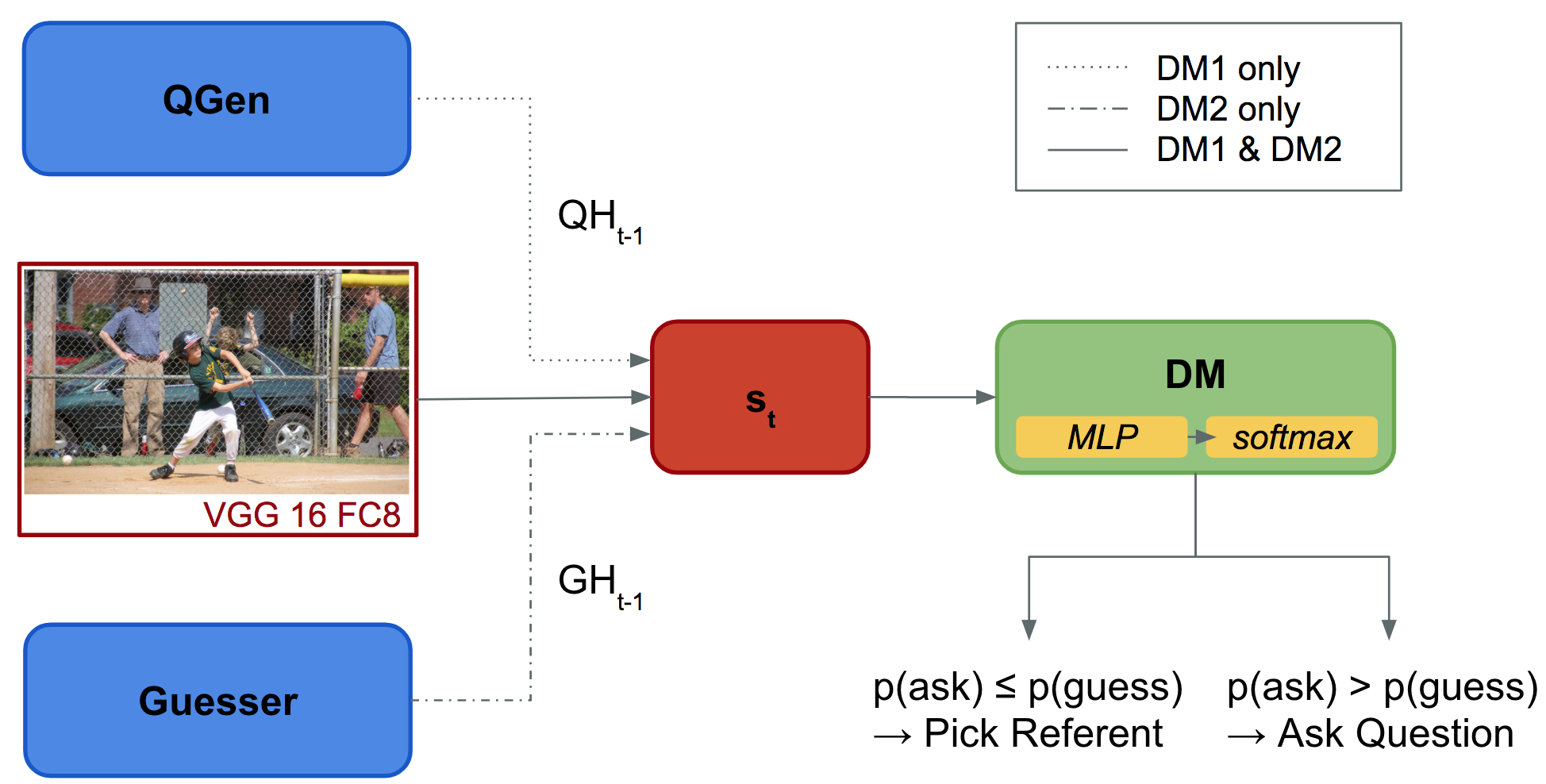} 
\caption{Our Questioner Model with two versions of the Decision Making module, DM1 and DM2. \label{deciders}}
\end{wrapfigure}

We experimented with two versions of the DM component, DM1 and DM2, 
which differ with respect to the encoding of the linguistic input they have access to.
Both decision makers have access to the image and implicitly to the
linguistic dialogue history: DM1 exploits the dialogue encoding learned by QGen's LSTM, which is trained to record information relevant for generating a follow-up question. 
In contrast, DM2 leverages the dialogue encoding learned by the Guesser's LSTM, which is trained to capture the properties of the linguistic input that are relevant to make a guess.
The two versions are illustrated in Figure~\ref{deciders}. 
The DM takes as input the concatenation ($s_t$) of the visual
features and of the dialogue history representation it gets either via QGen's
hidden state $QH_{t-1}$ (DM1) or the Guesser's hidden state $GH_{t-1}$ (DM2).
The resulting vector is passed through a MLP. The
output is then scaled between 0 and 1 by the Softmax function and
treated as a probability distribution.

We also implemented a hybrid DM module that receives as input
both $QH_{t-1}$ and $GH_{t-1}$, but we obtained worse performance.
Instead of insisting on the hybrid architecture (left for future
work), we maintain the two versions, DM1 and DM2, separated here so
that we can focus on investigating their differences and their
complementary contributions.\footnote{In principle, a decision-making
  module could also leverage the entropy of the scores for each
  candidate object produced by the Guesser. However, given the setup
  of the baseline Guesser (which we keep untouched for comparability),
  this would lead to implicitly using the symbolic object categories
  and spatial coordinates of each candidate object. This information,
  however, is not available to the humans at the time of deciding
  whether to guess or to continue asking: the list of candidate
  targets only becomes available once a participant decides to
  guess. Our DM modules only use the raw visual features, and thus are
  in a similar position to humans performing the task.}

\section{Experiments and Results}
\label{sec:experiments}

Next, we present our experimental setup and accuracy results. In Section~\ref{sec:analysis} we then
provide an in-depth analysis of the games and dialogues.

\subsection{Experimental setup}


The modules in the system by \newcite{guesswhat_game} (QGen, Guesser, and Oracle) are trained independently with supervised learning. To allow for direct comparison with their model, we follow the same setup for training the three original modules and also our new decision making module. Details on hyperparameter settings are provided in Appendix A. We use the same train, validation, and test sets as~\newcite{guesswhat_game}.

Both DM1 and DM2 are trained with Cross Entropy loss in a supervised manner,
which requires decision labels for the dialogue state after each question/answer pair. 
We use two different approaches to
obtain decision labels from the \emph{GuessWhat?!} ground truth
dialogues. In the first approach, the question/answer pairs are labelled based on the
human decision to {\em ask} or {\em guess}, obtained by checking whether there is  
a follow-up question in the human-human dialogues. We refer to
this paradigm as \emph{gt-label} (\emph{gt} for \emph{ground truth}). 
In the second approach, we label the question/answer
pairs based on whether the Guesser module is able to correctly predict
the target object given the current dialogue fragment.  
If the Guesser module is able to make a correct prediction after a given question/answer pair, 
we label that dialogue state with \emph{guess} and otherwise with \emph{ask}. 
This is referred to as \emph{guess-label}.  
DM1 can only be trained with \emph{gt-label} (since it does not have access to information coming from the Guesser). 
For DM2, we treat the choice between these two labelling approaches as a hyperparameter to be tuned on the validation set. 
DM2 achieves better results when trained with \emph{guess-label}. 
Experiments on the test set are then conducted with the optimal settings.

\subsection{Accuracy Results}

We first report results on task accuracy, i.e., the percentage of games where the task is successfully accomplished. 
Human accuracy is 90.8\%. Humans can ask as many questions as they like and, on average, they guess
after having asked 5.12 questions.
Table~\ref{tab:overallaccuracy} gives an overview of the accuracy results obtained by the baseline model, which always asks a fixed number of questions, and by our extended model with the two different versions of the DM, which decides when to ask or guess. 
We report the accuracy of our re-implementation of the baseline system, which is slightly better than the one reported by \newcite{guesswhat_game} for 5 questions.\footnote{\newcite{guesswhat_game} report an accuracy of 34\% with a fixed number of 5 questions, while 40.8\% is reported on the first author's GitHub page. Our re-implementation of the Oracle and
Guesser obtain an accuracy of 78.47\% (78.5) and 61.26\% (61.3), respectively (in brackets, the original accuracies reported by \newcite{guesswhat_game}).} 
To highlight the impact of including a DM module, we report the accuracies
the models achieve when changing the maximum number of questions allowed (MaxQ).  
When MaxQ = 5, 
the accuracies of the model enriched with a DM module are very similar to the baseline model 
(slightly lower for DM1: 40.45\%, and slightly higher for DM2: 42.19\%), 
and the average number of questions asked by the DM models is also comparable
to the 5 questions asked by the baseline, namely 4.62
(DM1) and 4.53 (DM2). However, if we observe how the model accuracy
varies when increasing MaxQ, we see that
the models equipped with a DM module tend to perform better than the baseline model
and that they do so by asking fewer questions than the baseline on average. 
Furthermore, of the two models enriched with a decision maker, DM1 is more stable across the
various settings both in terms of accuracy and of number of questions asked.

\begin{table}\small
\begin{center}
\begin{tabular}{|l|l|l|l|l|l|l|l|}
\hline
MaxQ			& 5 & 8 & 10 & 12 & 20 & 25 &30\\\hline\hline
Baseline 	& 41.18 & 40.7  & 39.8  & 38.96& 36.02& 35.918 & 35.88\\ \hline
DM1  & 			40.45 (4.62) 	&  40.12 (6.17) & 40.02 (6.70)  &40.00 (6.97)   &39.87 (7.14) &39.68 (7.14) &39.68 (7.14)\\ \hline
DM2 	& 42.19 (4.53)  & 41.30 (7.22) & 41.12 (8.71) & 39.73 (10.72) & 37.75 (13.39) & 36.86 (13.47) & 36.83(13.51) \\
\hline
\end{tabular}
\caption{Accuracy on the entire test set (all games, viz. including
successful, unsuccessful, decided and undecided ones) for task success by varying
  the maximum number of questions allowed (MaxQ). Within brackets, the average
number of questions asked for each setting. The baseline model always asks the maximum number of questions allowed.
\label{tab:overallaccuracy}}
\end{center}
\end{table}

In the following analysis, unless indicated otherwise, we consider the baseline model with 5 questions, because it yields the highest baseline accuracy, and the DM models with MaxQ = 10, because more than 90\% of the games solved by humans contain up to 10 questions.


\section{Analysis}
\label{sec:analysis}

To better understand the results reported above and to gain insight on how the inclusion of a decision making component affects the resulting dialogues, we carry out an analysis of the behaviour of our models. We first examine how the complexity of the game (as determined by visual properties of the image) affects performance, and then  analyse the quality of the linguistic interaction from the perspective of the Questioner role. 

\subsection{Complexity of the Game}
\label{sec:complexity}

Intuitively, the more complex the image involved in a round of the game, the harder it is to guess the target
object. As a proxy for image complexity, we consider the following measures:
(i) the number of objects in the image, (ii) the number of objects
with the same category as the target object, and (iii) the size of the
target object, which we compute in terms of the proportion of the
cropped target object area with respect to the whole image. The
distribution of games in the whole dataset is fairly balanced across
these factors. See Appendix B for full details. 
  
We fit a linear logistic regression model for the task of predicting
whether a game will be successful or unsuccessful (i.e., whether the
right target object will be selected), using the three measures above
as independent predictor variables. It should be noted that in some
cases our Questioner model may reach the maximum number of 10
questions without ever taking the action to guess. We refer to these
games as \emph{undecided}, whereas we call \emph{decided} games those
games where the DM lets the Guesser pick a referent within the maximum
number of questions allowed. Out of the whole test set, the
  amount of decided games is 77.67\% and 15.58\% for DM1 and DM2,
  respectively.\footnote{To understand the rather big difference
    between the number of decided games in DM1 and DM2, we also
  evaluated the models using ground truth data for the QGen and
  Oracle modules. When human-human dialogues are used as input, the
  percentage of decided games is high and virtually identical for the two versions of the DM module (81.31\% for DM1 and 81.30 for DM2.) This shows that DM2 is affected much more than DM1 by the errors produced by other
  modules. We leave an analysis of this aspect to future work. Throughout the present paper, all results and analyses reported are not based on ground truth data, but on the representations automatically generated by other modules.} 
It is critical to take this into account, since cases where the model is
simply forced to make a guess are less informative about the
performance of the DM module. Therefore, we fit two additional
logistic regression models using the same three predictors: one where
we consider only decided games and predict whether they are
\emph{successful} or \emph{unsuccessful}, and one where the dependent
variable to be predicted are the \emph{decided} vs.~\emph{undecided}
status of a game.

In Table~\ref{tab:regression}, we report the regression coefficients for the different logistic regression models, estimated with iteratively reweighted least squares.\footnote{We use the R implementation of the logistic regression algorithm.} 
Plots of the predictor variables are available in Appendix B.
Regarding the distinction between successful and unsuccessful games, we observe that the three image complexity measures we consider play a significant role. The three models (baseline, DM1, and DM2) are more likely to succeed in games that are intuitively easier --- i.e., when the image contains fewer objects overall and fewer objects of the same category as the target (negative coefficients), and when the relative size of the target object is larger (positive coefficients). 
Interestingly, when we look into the distinction between decided and undecided games, we observe different behaviour for the two versions of the DM module. While DM1 tends to make a decision to stop asking questions and guess in easier games, surprisingly DM2 is more likely to make a guessing decision when the image complexity is higher (note the contrasting tendency of the coefficients in the last column of Table~\ref{tab:regression} for DM2). However, similarly to DM1, once DM2 decides to guess (decided games in Table~\ref{tab:regression}), the simpler the image the more likely the model is to succeed in picking up the right target object.

\begin{table}
\begin{center}\small
\begin{tabular}{|l||rrr|rr||rr|}\hline
 & \multicolumn{5}{|c||}{\emph{successful} vs.~\emph{unsuccessful}} & \multicolumn{2}{c|}{\emph{decided} vs.~\emph{undecided}} \\ \hline\hline
 & \multicolumn{3}{|c|}{all games} &  \multicolumn{2}{c||}{decided games} & \multicolumn{2}{c|}{all games} \\\hline
 & Baseline & DM1 & DM2 & DM1 & DM2 & DM1 & DM2 \\\hline
  \# objects & -0.213094 & -0.212920 & -0.217468 & -0.220929 & -0.23967 & -0.05292 &  0.144233  \\        
 \# objects same cat. as target&-0.150294 & -0.144740 &-0.150090 &-0.148251 & -0.165415 & -0.058392  & 0.087068 \\
\% target object's area & 4.88720&4.254 & 3.82114 & 4.15606& 7.0496 & 1.59634 & -2.38053 \\\hline
\end{tabular}
\end{center}
\caption{Estimated regression coefficients for the logistic regression models distinguishing between \emph{successful} vs.~\emph{unsuccessful} and \emph{decided} vs.~\emph{undecided} games. A positive/negative coefficient indicates a positive/negative correlation between a predictor and the probability of the \emph{successful} or \emph{decided} class. The contribution of all predictors is statistically significant in all models ($p < 0.0001$).\label{tab:regression}}
\end{table}

\subsection{Quality of the Dialogues}
\label{sec:quality}

As stated in the introduction, we believe that a good visual dialogue
model should not only be measured in terms of its task success but
also with respect to the quality of its dialogues. In particular
unnatural repetitions and unnecessary questions should be avoided.
Hence, here we look into the dialogues produced by the different Questioner models comparing them with
respect to these two criteria.

\paragraph{Repeated questions}
Qualitative examination of the dialogues shows that the Questioner
models often ask the same question over and over again. This results
in unnatural linguistic interactions that come across as incoherent
--- see, for example, the dialogue in Figure~\ref{fig:qualitative}
(bottom part). We analyse the dialogues produced by the models in
terms of the amount of repetitions they contain. For the sake of
simplicity, we only consider repeated questions that are exact string
matchings (i.e., verbatim repetitions of entire
questions).  In this case, we consider the baseline
model that asks 10 questions, as this makes for a fairer comparison
with our models, where MaxQ = 10 (see end of
Section~\ref{sec:experiments}).

Table~\ref{tab:repetition} reports the percentage of
dialogues that contain at least one repeated question (across-games)
and the percentage of repeated questions within a dialogue averaged
across games (within-game). We check the percentages with respect to
both all the games and only decided games. When considering all games, we
find that the baseline model produces many more dialogues with
repeated questions (98.07\% vs.~74.66\% for DM1 and 84.05\% for DM2)
and many more repeated questions per dialogue (45.74\% vs.~23.34\% for
DM1 and 39.27\%~for DM2) than our DM models. Among our models, the
dialogues by DM1 are less repetitive than those by DM2. However, the
difference between the two DM models is reversed when zooming into the decided
games.  

Our method for quantifying repeated questions is clearly simplistic: questions that have an identical surface form may not count as mere repetitions if they contain pronouns that have different antecedents. For example, a question such as ``Is
it the one on the left?'' could be asked twice within the same dialogue with different antecedents for the anaphoric phrase ``the one''. In contrast, several instances of a question that includes a noun referring to a candidate object (such as ``Is it a dog?'') most probably are true repetitions that should be avoided. Therefore, as a sanity check, we perform our analysis taking into account only questions that mention a candidate object.\footnote{A list of objects is provided in Appendix C.} As shown in Table~\ref{tab:repetition}, this yields the same patterns observed when considering all types of questions. 

%
%

\begin{table}
\small
\begin{center}
\begin{tabular}{|@{\ }r@{\ }|c|c|c|c|c|c|c|c|c|c@{}|}
\hline
& \multicolumn{6}{@{}c}{All games} & \multicolumn{4}{|c|}{Decided games}\\\hline\hline
 & \multicolumn{3}{@{}c}{Overall} & \multicolumn{3}{|c|}{Objects} & \multicolumn{2}{c}{Overall} & \multicolumn{2}{|c|}{Objects} \\\hline
			& Baseline & DM1 & DM2 & Baseline & DM1 & DM2
                                   & DM1 & DM2 & DM1 & DM2 \\\hline
across-games 	& 98.07\% &  74.66\%  & 84.05\% & 44.88\%& 32.94\%& 40.10\%
                                   & 69.18\% & 7.90\% & 67.05\% & 7.79\%\\ \hline
within-game  & 45.74\% 	&  23.34\% & 39.27\% & 18.38\% & 11.61\%& 16.42\% &
                                                                      31.28\%
                                         & 12.52\% & 30.06\%& 12.36\%\\ \hline
\end{tabular}
\caption{Percentages of repeated questions in all games and in decided games. Overall: all types of questions; Objects: only questions mentioning a candidate target object. All differences between the baseline and our models are statistically significant (Welch $t$-test with $p < 0.0001$).}
\label{tab:repetition}
\end{center}
\end{table}

\paragraph{Unnecessary questions}
Another feature of the dialogues revealed by qualitative examination
is the presence of questions that are not repetitions of earlier
questions but that, in principle, are not needed to successfully solve
the game given the information gathered so far. For example, the last
three questions in the dialogue in Figure~\ref{fig:needofus} in the
Introduction are not necessary to solve the game, given the evidence
provided by the first two questions.
By including a decision making component, our models may be able to alleviate this problem. 
In this analysis, we compare the baseline system that asks 5 questions to our models with MaxQ = 10 and look into cases where our models ask either fewer or more questions than the baseline.

Table~\ref{tab:state} provides an overview of the results. 
When considering all the games, we see
that the DM models ask many more questions (64.43\% DM1 and 85.14\%
DM2) than the baseline. This is not surprising, since many games are undecided (see Section~\ref{sec:complexity}) and hence contain more questions than the baseline (10 vs.~5). 
Zooming into decided games thus allows for a more appropriate comparison. 
Table~\ref{tab:state} also includes information on whether  
asking fewer or more questions helps (+~Change),
hurts (--~Change) or does not have an impact on task success (No Change) with respect to the baseline results. 
We observe that DM2 dramatically decreases the number of questions:
in 95.17\% of decided games, it asks fewer questions than the baseline;
interestingly, in only 13.98\% of cases where it asks fewer
questions its performance is worse than the baseline --- in all the other
cases, either it achieves the same success (56.18\%) or even improves on the
baseline results (25.01\%). The latter shows that DM2 is able to reduce the number of
unnecessary questions, as illustrated in Figure~\ref{fig:needofus}. On the other hand, DM1 does not seem to reduce the
number of unnecessary questions in a significant way.


\begin{table}
\begin{center}
\small
\begin{tabular}{|c|c|c|c|c|c|c|c|c||c|c|}
  \hline
 & \multicolumn{8}{c||}{Decided games} & \multicolumn{2}{c|}{All
                                   games}\\\hline \hline
  \multirow{2}{*}{DM}
      & \multicolumn{2}{c|}{+ Change}
          & \multicolumn{2}{|c|}{-- Change}
          & \multicolumn{2}{|c|}{No Change}
          & \multicolumn{2}{|c||}{Total} & \multicolumn{2}{|c|}{Total}
          \\             \cline{2-11}
  & Fewer & More  & Fewer & More & Fewer & More &Fewer & More & Fewer
                  & More  \\  \hline
  DM1 & 1.77 & 3.46 & 2.64 & 3.79  & 22.58 & 50.35 & 26.99 & 57.6  &
                                                                     22.63
                  & 64.43
  \\      \hline
  DM2 & 25.01 & 0.16  & 13.98 & 0.81  & 56.18 & 3.67 & 95.17 & 4.64 &
                  14.83 & 85.14\\      \hline
\end{tabular}
\caption{Games played by DM with MaxQ=10, and the baseline with 5
  fixed questions. Percentages of games (among all games and only decided games)
  where the DM models ask either fewer or more questions than the
  baseline.  For the decided games, percentages of
  games where asking fewer/more questions helps (+ Change), hurts (-- Change) or does not have an impact on task success w.r.t. the baseline result (No Change).}
\label{tab:state}
\end{center}
\end{table}

\subsection{Discussion}

\begin{figure}\centering
\includegraphics[width=15cm]{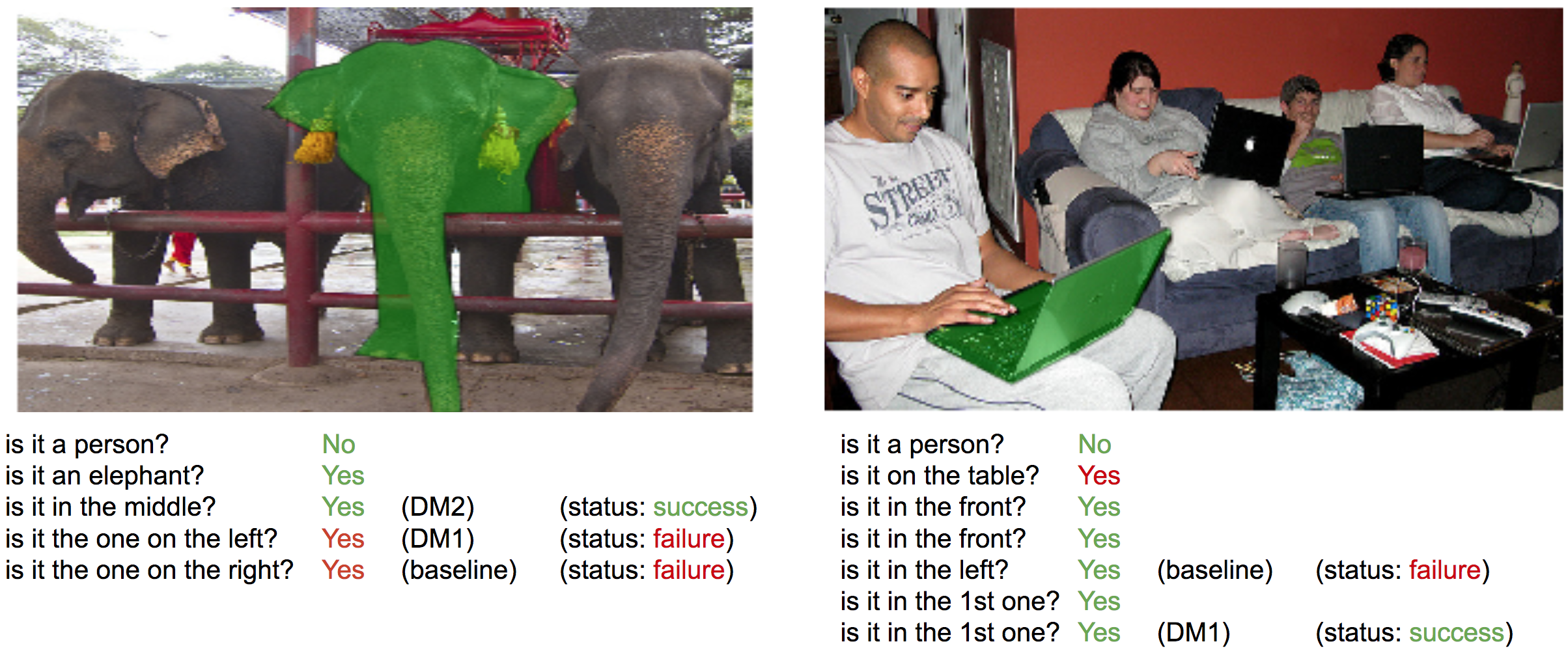}
\caption{Examples where our model achieves task success by asking fewer or more questions than the baseline. Answers in red highlight Oracle errors. QGen often produces repeated or incoherent questions.}
\label{fig:qualitative}
\end{figure}


Our analyses show that using a decision making component produces dialogues with fewer repeated questions and can reduce the number of unnecessary questions, thus potentially leading to more efficient and less unnatural interactions. Indeed, for some games not correctly resolved by the baseline system, our model is able to guess the right target object by asking fewer questions. DM2 is substantially better at this than DM1 (25.01\% vs.~1.77\% of decided games; see Table~\ref{tab:state}). 
By being restricted to a fixed number of questions, the baseline system often introduces noise or apparently forgets about important information that was obtained with the initial questions. Thanks to the DM component, our model can decide to stop the dialogue once there is enough information and make a guess at an earlier time, thus avoiding possible noise introduced by Oracle errors, as illustrated in Figure~\ref{fig:qualitative} (left). Qualitative error analysis, however, also shows cases where the DM makes a premature decision to stop asking questions before obtaining enough information. Yet in other occasions, the DM seems to have made a sensible decision, but the inaccuracy of the Oracle or the Guesser components lead to task failure. Further examples are available in Appendix D.

In some games with complex images, the information obtained with 5 questions (as asked by the baseline) is not enough to resolve the target. The flexibility introduced by the DM allows our model to reach task success by asking additional questions. Figure~\ref{fig:qualitative} (right) gives an example. Furthermore, if noise has been introduced at earlier stages of the dialogue, asking further questions can increase the chance to recover relevant information.
However, we also observe that in some games correctly guessed by the baseline system, asking more questions leads to failure since it opens the door to getting wrong information from the Oracle. 

In the current analyses, we have not studied the behaviour of our DM models when using ground truth data instead of the noisy automatic output produced by the other modules.  In part, this is motivated by our long-term goal of developing fully data-driven multimodal conversational agents that can be trained end-to-end. We leave for future work carrying out a proper ablation study that analyses the impact of using ground truth vs.~automatic data on the DM component.

\section{Conclusion}
\label{sec:conclusion}

Research on dialogue systems within the Computational Linguistics community has shown the importance of equipping
such systems with dialogue management capabilities.
Computer Vision
researchers have launched the intriguing Visual Dialogue challenge
mostly focusing on comparing strong machine learning paradigms on task accuracy, and largely ignoring the aforementioned line of research on dialogue systems.  Our
goal is to explore how data-driven conversational agents, modelled by
neural networks without additional annotations usually exploited by traditional dialogue systems, can profit from a dialogue
management module. The present work is a first step towards this
long-standing goal. 
We have taken the \emph{GuessWhat?!}~task as our testbed, since it provides a simple setting with elementary question-answer sequences and is task-oriented, which opens the door to using an unsupervised approach in the future.   
We have focused on augmenting the Questioner agent of the
\emph{GuessWhat?!}~baseline, which consists of a Question Generator
and a Guesser module, with a decision making component (DM) that determines
after each question-answer pair whether the Question Generator should
ask another question or whether the Guesser should guess the target
object. 
The solution we propose is technically simple, and we believe promising and more cognitive
principled than, for example, including a \texttt{stop} token as~\newcite{stru:end17}.  
We show that incorporating a decision making
component does lead to less unnatural dialogues. It remains
to be seen whether a hybrid DM module that exploits the dialogue encodings of both the Question Generator and
the Guesser modules could bring further qualitative and quantitative
improvements.

\cut{\eb{The decision module conditions its decisions directly on the
  internal states of the other modules (QGen or Guesser), making the
  architecture self-contained and end-to-end trainable. We can
  consider the decider module as a way to automatically
  introspect/monitor our model and understand when enough evidence to
  make a guess has been cumulated by the system. This is only possible
  because it bases its decisions on the internal states of the other
  modules, without any external information. In future work we will be
  able to exploit the deciding module to induce more strategic
  questions, for example toward accelerating the decision process.}
}

\section*{Acknowledgements}
 We kindly acknowledge the European Network on Integrating Vision and
 Language (iV\&L Net) ICT COST Action IC1307. The Amsterdam team is partially funded by the Netherlands Organisation for Scientific Research (NWO) under VIDI grant nr.~276-89-008, {\em Asymmetry in Conversation}.
 In addition, we gratefully acknowledge the support of NVIDIA
 Corporation with the donation to the University of Trento of the
 GPUs used in our research.

\bibliographystyle{acl}
\bibliography{raffa,raq,tim}

\section*{Appendix A: Details of {\em GuessWhat?!} Dataset and Experimental  Setup}

\paragraph{Dataset.}
The GuessWhat?! dataset contains 77,973 images with 609,543 objects
and around 155K human-human dialogues. The dialogues contain around
821K question/answer pairs composed out of 4900 words (counting
  only words that occur at least 3) on 66,537 unique images
and 134,073 target objects. Answers are Yes (52.2\%), No
(45.6\%) and NA (not applicable, 2.2\%); dialogues contain on average
5.2 questions and there are on average 2.3 dialogues per image.  There
are successful (84.6\%), unsuccessful (8.4\%) and not completed
(7.0\%) dialogues.

\paragraph{Games and Experimental Setting.}
In the baseline model, games consist of 5 turns each consisting of question/answer pairs asked by QGen
and answered by the Oracle. The QGen module stops asking questions
after having received the answer to the 5th question. It is then the
turn of the Guesser.

All the modules are trained independently using Ground Truth data.
For the visual features, `fc8' of the VGG-16 network is used. Before
visual feature calculation, all images are resized to 224X224.
For each object, the module receives the representation of the object category, viz., a dense category embedding
obtained from its one-hot class vector using a learned look-up table,
and its spatial representation, viz., an 8-dimensional vector.
The dialogue is encoded using variable length LSTM with 512 hidden size for Guesser and
Oracle and 1024 hidden size for QGen. 
The LSTM, object category/word look-up tables and MLP parameters are
optimized while training by
minimizing the negative log-likelihood of the correct answer using
ADAM optimizer with learning rate 0.001 for Guesser and Oracle. For QGen,
 the conditional log-likelihood is maximized based on the next question 
 given the image and dialogue history. All the parameters are
tuned on the validation set, training is stopped when there is no
improvement in the validation loss for 5 consecutive epochs and best
epoch is taken.

\section*{Appendix B: Analysis Regarding Image Complexity}

\begin{figure*}[ht]
\hspace*{-0.5cm}
\subfloat[\# instances of target object.]{
	\label{fig:datasetinstance}
	\begin{minipage}{5.4cm}
	\includegraphics[width=6cm]{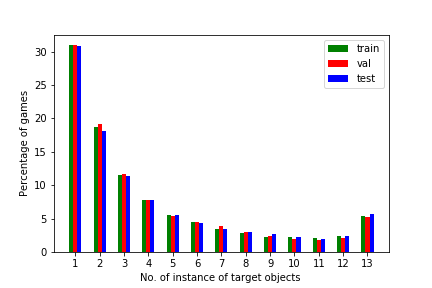} 
	\end{minipage}
}\hspace*{-0.2cm}
\subfloat[\# of objects.]{
	\label{fig:datasetobject}
	\begin{minipage}{5.4cm}
	\includegraphics[width=6cm]{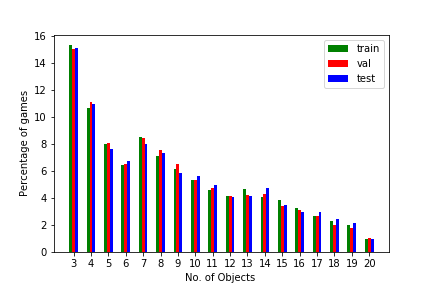} 
	\end{minipage}
}\hspace*{-0.2cm}
\subfloat[\% area covered by target object.]{
	\label{fig:datasetarea}
	\begin{minipage}{5.4cm}
	\includegraphics[width=6cm]{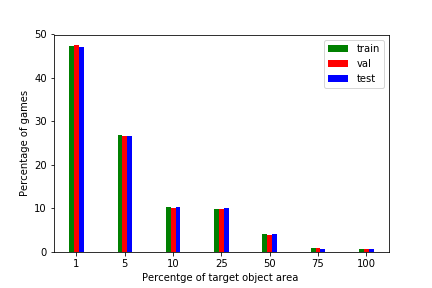} 
	\end{minipage}
}
\caption{Image distribution with respect to the image complexity measures in the different dataset splits.}
\label{fig:dist}
\end{figure*}

\begin{figure*}[ht]
\hspace*{-0.5cm}
\subfloat[no. of instances of target object.]{
	\label{fig:humaninstance}
	\begin{minipage}{5.3cm}
	\includegraphics[width=6cm]{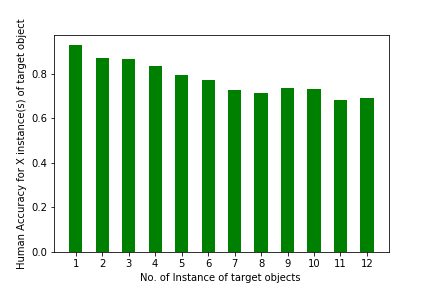} 
	\end{minipage}
}\hspace*{-0.1cm}
\subfloat[no. of objects.]{
	\label{fig:humanobject}
	\begin{minipage}{5.3cm}
	\includegraphics[width=6cm]{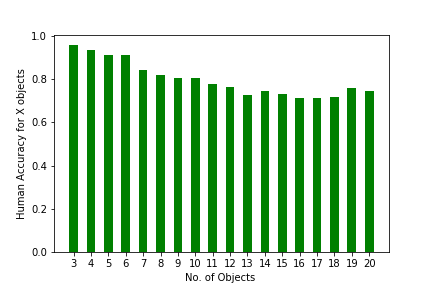} 
	\end{minipage}
}\hspace*{-0.2cm}
\subfloat[\% area covered by target object.]{
	\label{fig:humanarea}
	\begin{minipage}{5.3cm}
	\includegraphics[width=6cm]{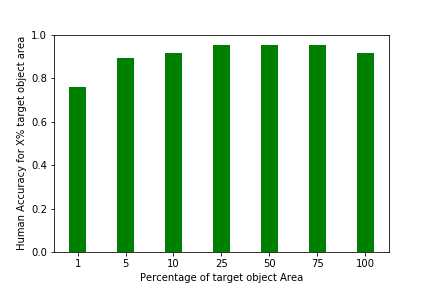} 
	\end{minipage}
}
\caption{Human accuracy distribution with respect to the image complexity features.}
\label{fig:human-analysis}
\end{figure*}

\paragraph{Distribution of image complexity measures.}
Figure~\ref{fig:dist} shows the image distribution across the train,
validation and test sets with respect to the image complexity
measures, namely (a) the number of instances of the target object, (b) the 
number of objects, and (c) the percentage of target object area with
respect to the overall image. We can see that the distribution with respect to these measures is very similar in the three sets. 
Figure~\ref{fig:human-analysis} provides human performance based on the different image complexity
measures. Human performance is similar to the model performance. While
human accuracy is comparatively high, it also decreases when the image complexity increases.

\paragraph{Image complexity measures as predictors in the logistic regression models.}

\begin{figure}[ht]
\subfloat[no. of instances of target object.]{
	\label{fig:dm1SvUinstance}
	\begin{minipage}{4.7cm}
	\includegraphics[width=4.7cm]{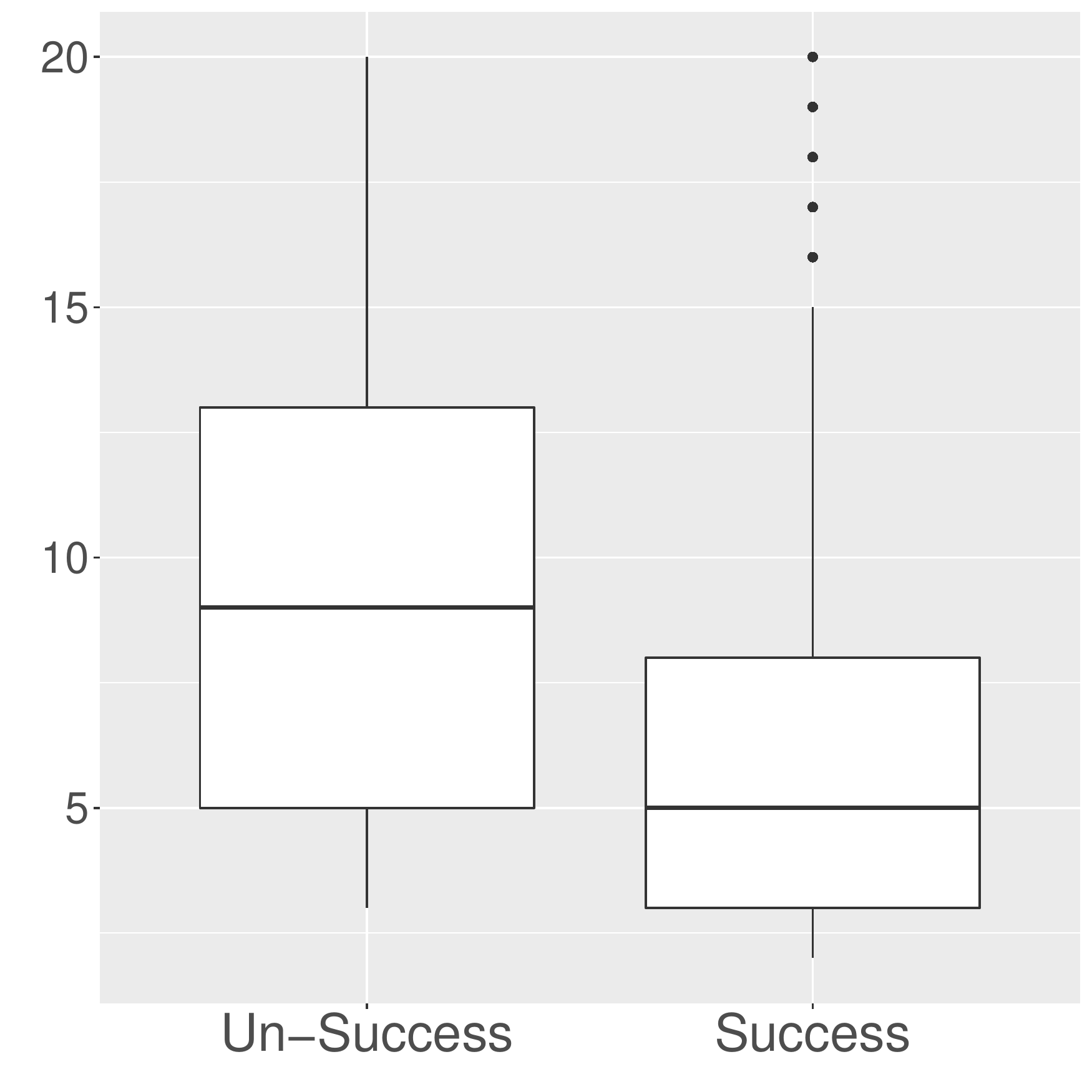} 
	\end{minipage}
}
\subfloat[no. of objects.]{
	\label{fig:dm1SvUobject}
	\begin{minipage}{4.7cm}
	\includegraphics[width=4.7cm]{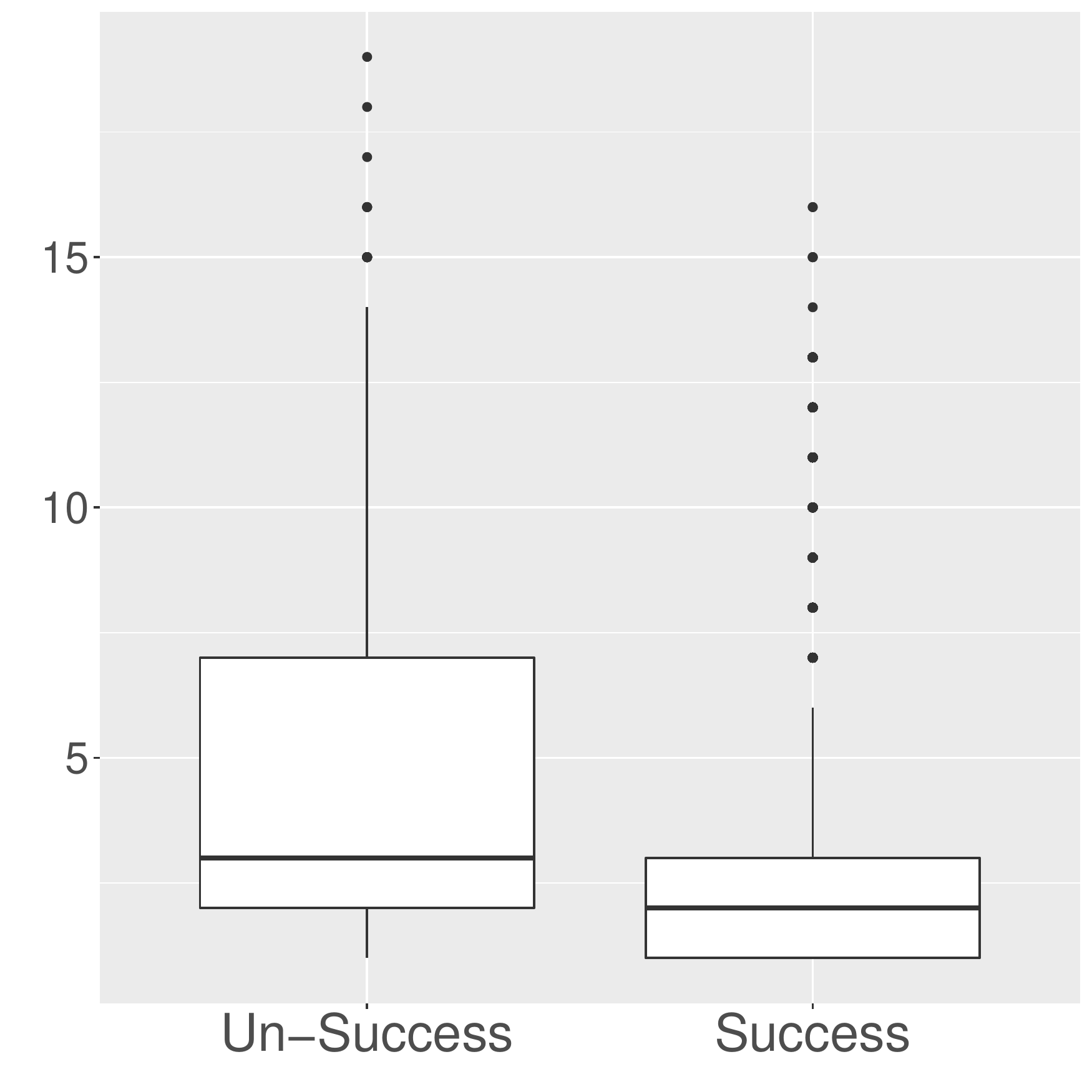} 
	\end{minipage}
}
\subfloat[\% area covered by target object.]{
	\label{fig:dm1SvUarea}
	\begin{minipage}{4.7cm}
	\includegraphics[width=4.7cm]{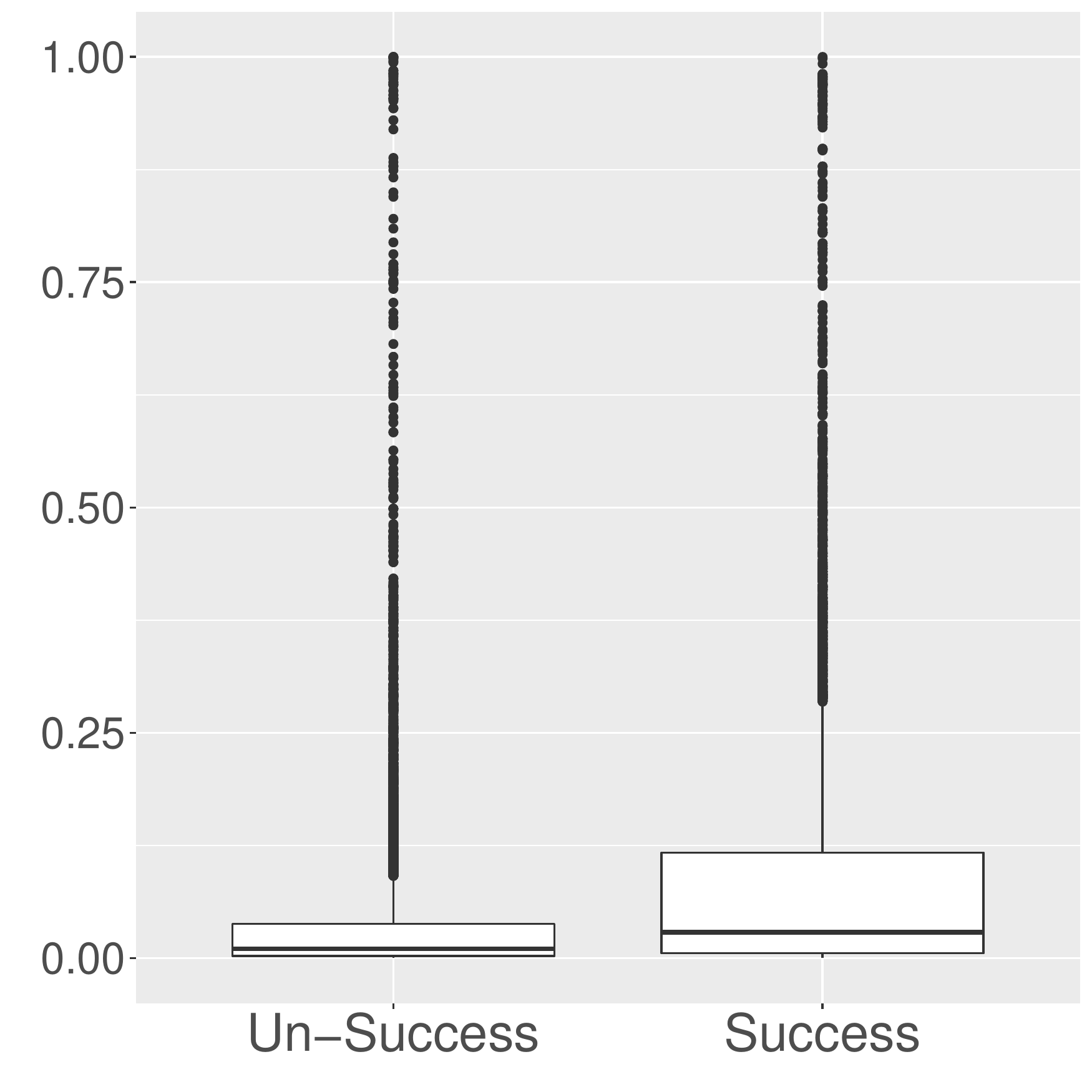} 
	\end{minipage}
}
\vspace*{-.2cm}
\caption{Effect of image complexity measures on successful vs.~unsuccessful
  games played by DM1.}
\label{fig:dm1SvU}
\vspace*{-.8cm}
\end{figure}

\begin{figure}[ht]
\subfloat[\# of instances of target object.]{
	\label{fig:dm2SvUinstance}
	\begin{minipage}{4.7cm}
	\includegraphics[width=4.7cm]{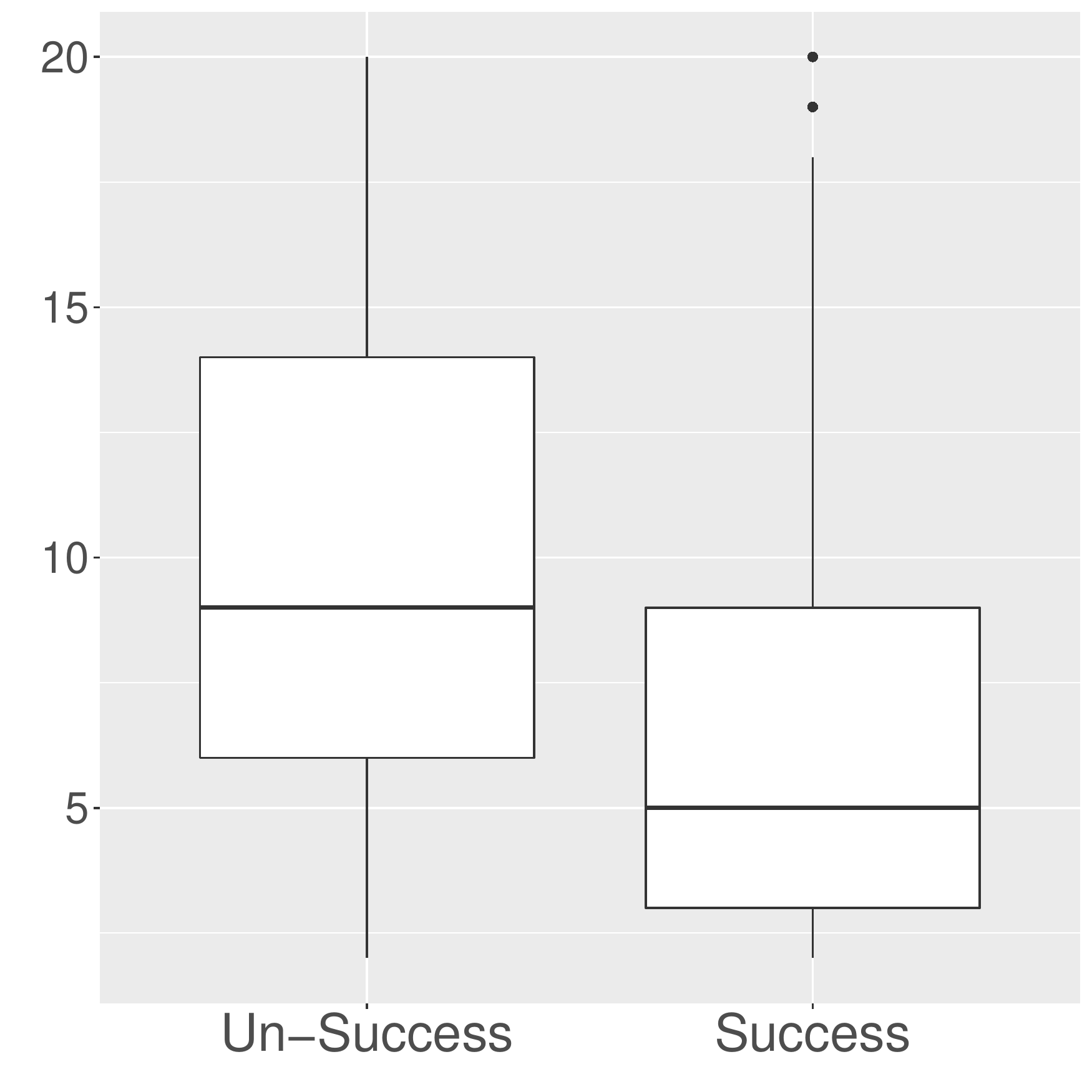} 
	\end{minipage}
}
\subfloat[\# of objects.]{
	\label{fig:dm2SvUobject}
	\begin{minipage}{4.7cm}
	\includegraphics[width=4.7cm]{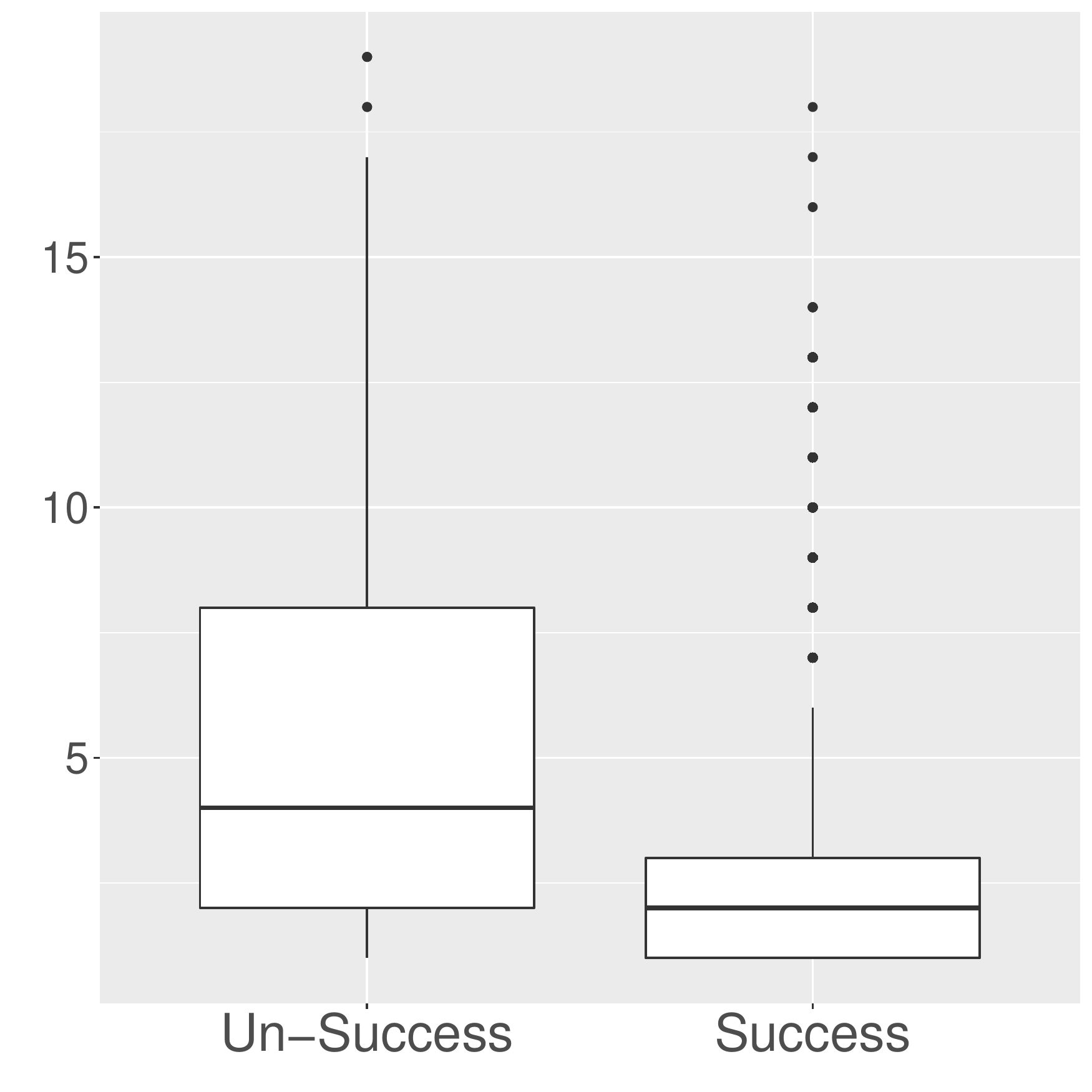} 
	\end{minipage}
}
\subfloat[\% area covered by target object.]{
	\label{fig:dm2SvUarea}
	\begin{minipage}{4.7cm}
	\includegraphics[width=4.7cm]{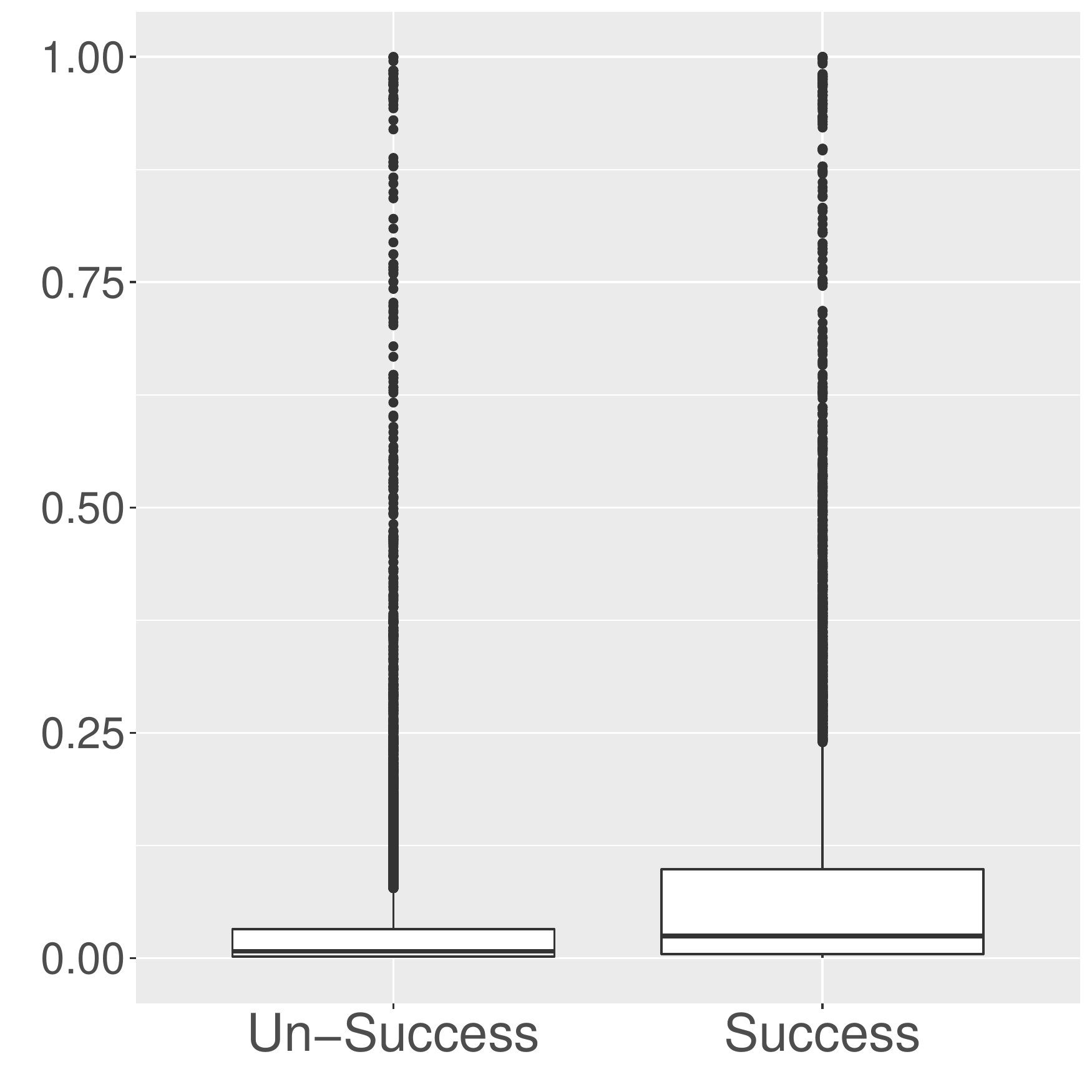} 
	\end{minipage}
}
\vspace*{-.2cm}
\caption{Effect of image complexity measures on successful vs.~unsuccessful
  games played by DM2.}
\label{fig:dm2SvU}
\vspace*{-.8cm}
\end{figure}

\begin{figure*}[ht]
\subfloat[\# of instances of target object.]{
	\label{fig:dm1DSvUinstance}
	\begin{minipage}{4.7cm}
	\includegraphics[width=4.7cm]{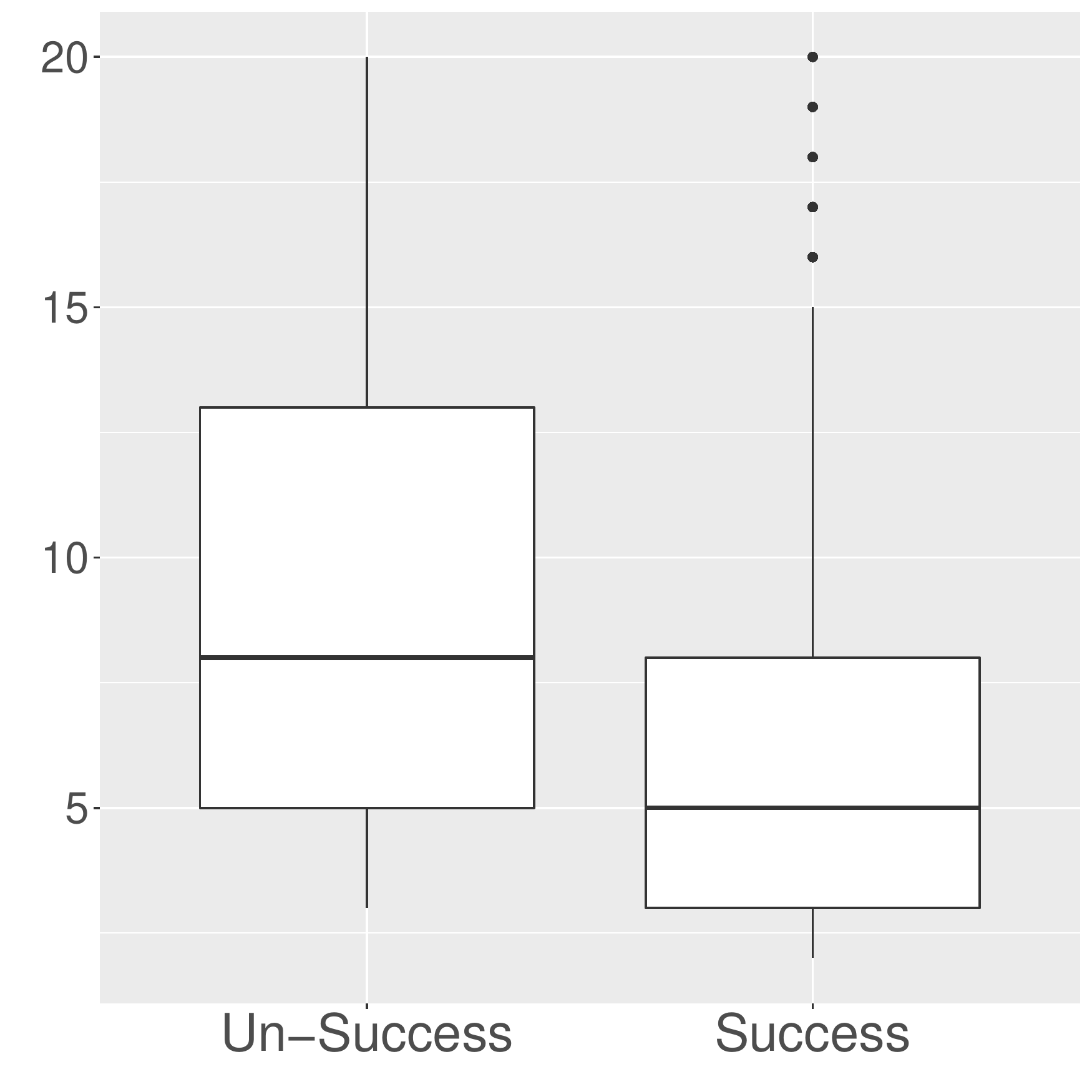} 
	\end{minipage}
}
\subfloat[\# of objects.]{
	\label{fig:dm1DSvUobject}
	\begin{minipage}{4.7cm}
	\includegraphics[width=4.7cm]{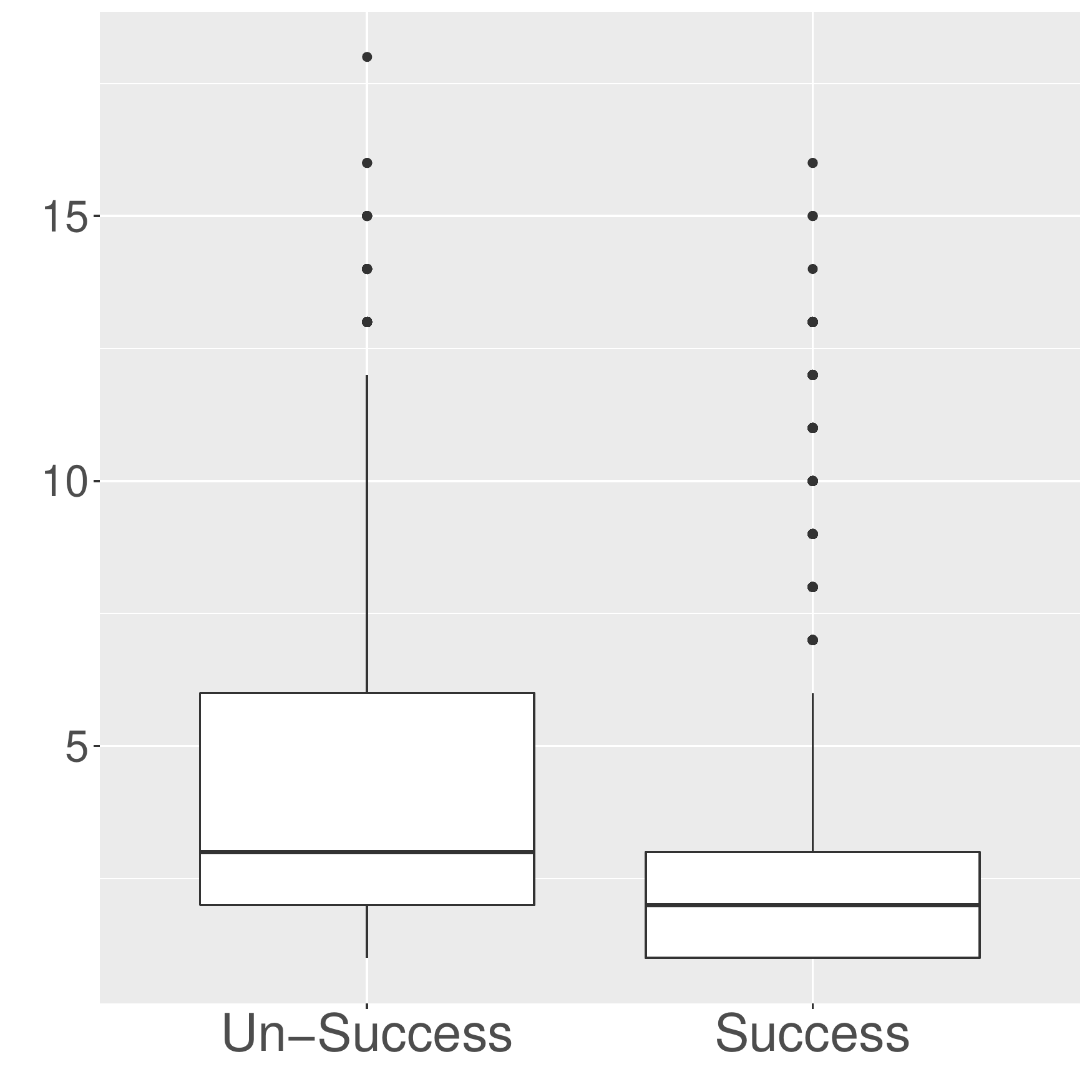} 
	\end{minipage}
}
\subfloat[\% area covered by target object.]{
	\label{fig:dm1DSvUarea}
	\begin{minipage}{5.3cm}
	\includegraphics[width=4.7cm]{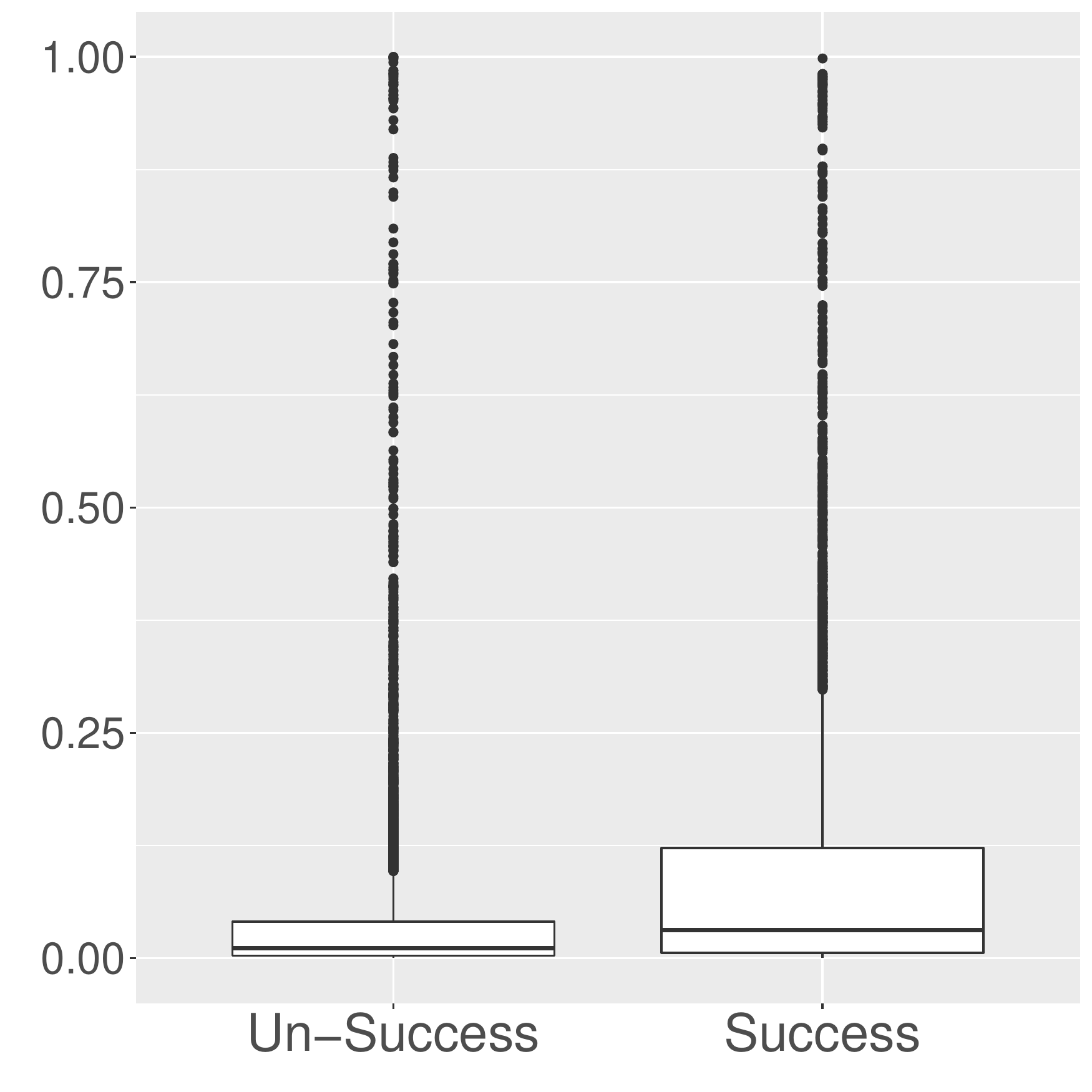} 
	\end{minipage}
}
\vspace*{-.2cm}
\caption{Effect of image complexity measures on successful vs.~unsuccessful {\em decided games} by DM1.}
\label{fig:dm1DSvU}
\vspace*{-.8cm}
\end{figure*}

\begin{figure*}[ht]
\subfloat[\# of instances of target object.]{
	\label{fig:dm2DSvUinstance}
	\begin{minipage}{4.7cm}
	\includegraphics[width=4.7cm]{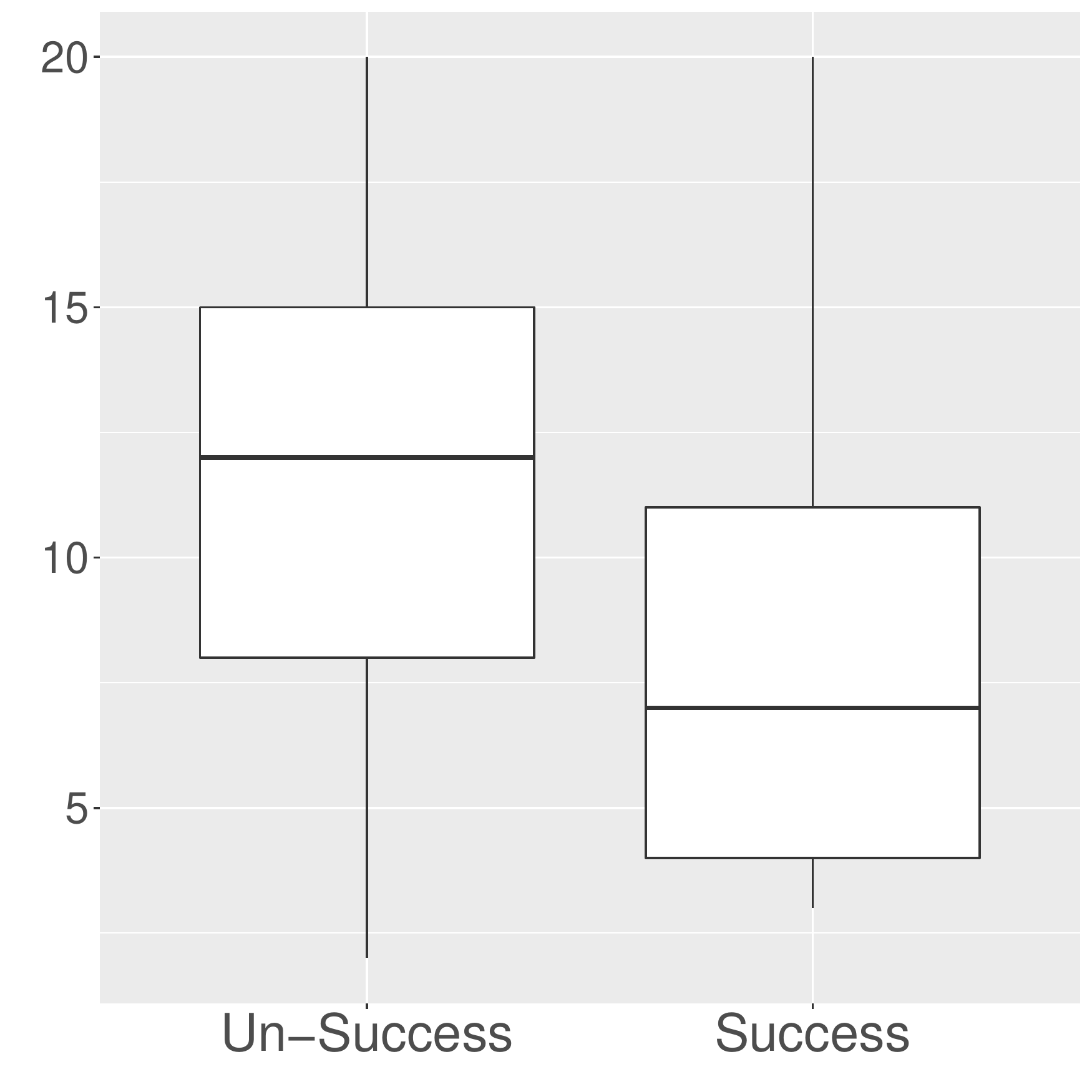} 
	\end{minipage}
}
\subfloat[\# of objects.]{
	\label{fig:dm2DSvUobject}
	\begin{minipage}{4.7cm}
	\includegraphics[width=4.7cm]{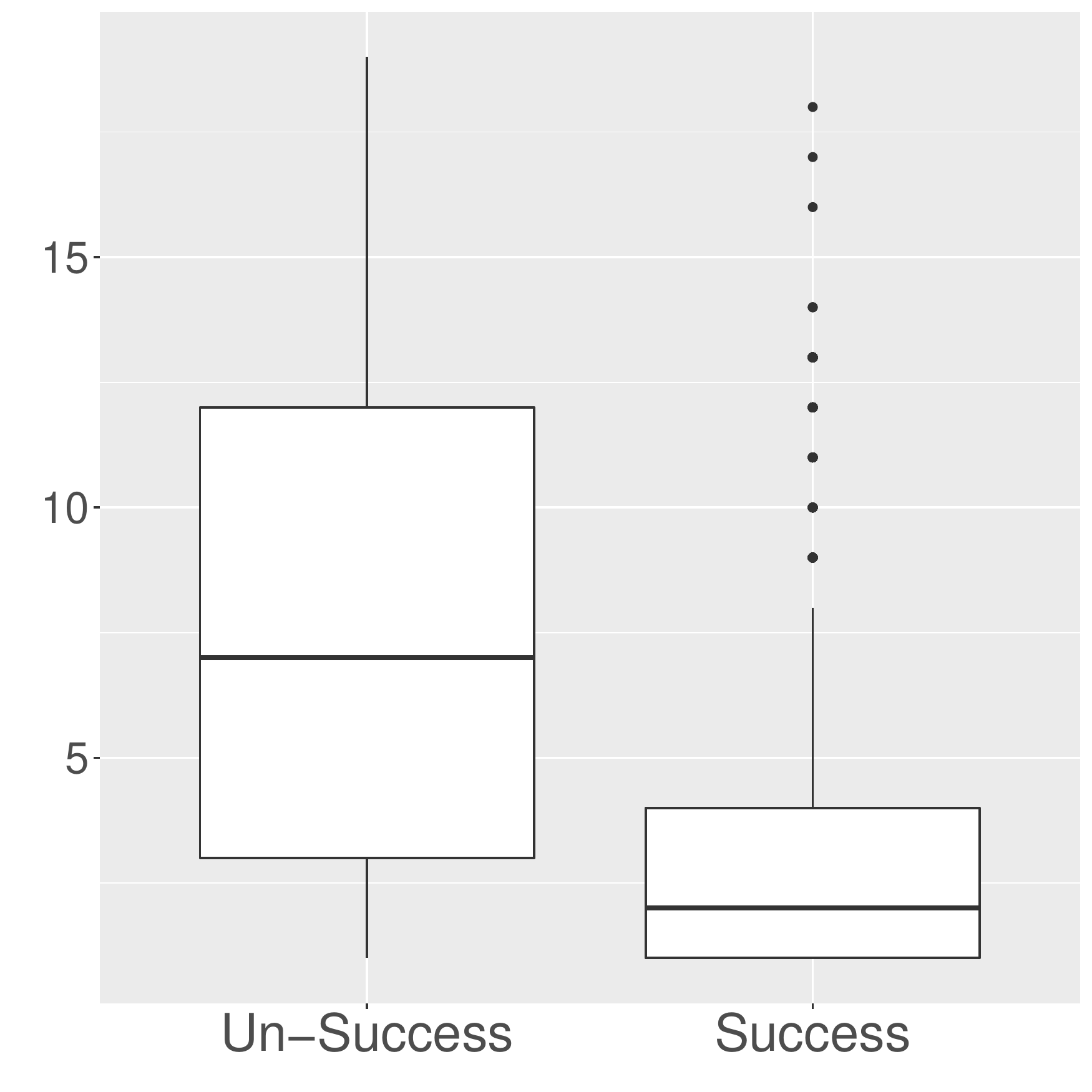} 
	\end{minipage}
}
\subfloat[\% area covered by target object.]{
	\label{fig:dm2DSvUarea}
	\begin{minipage}{4.7cm}
	\includegraphics[width=4.7cm]{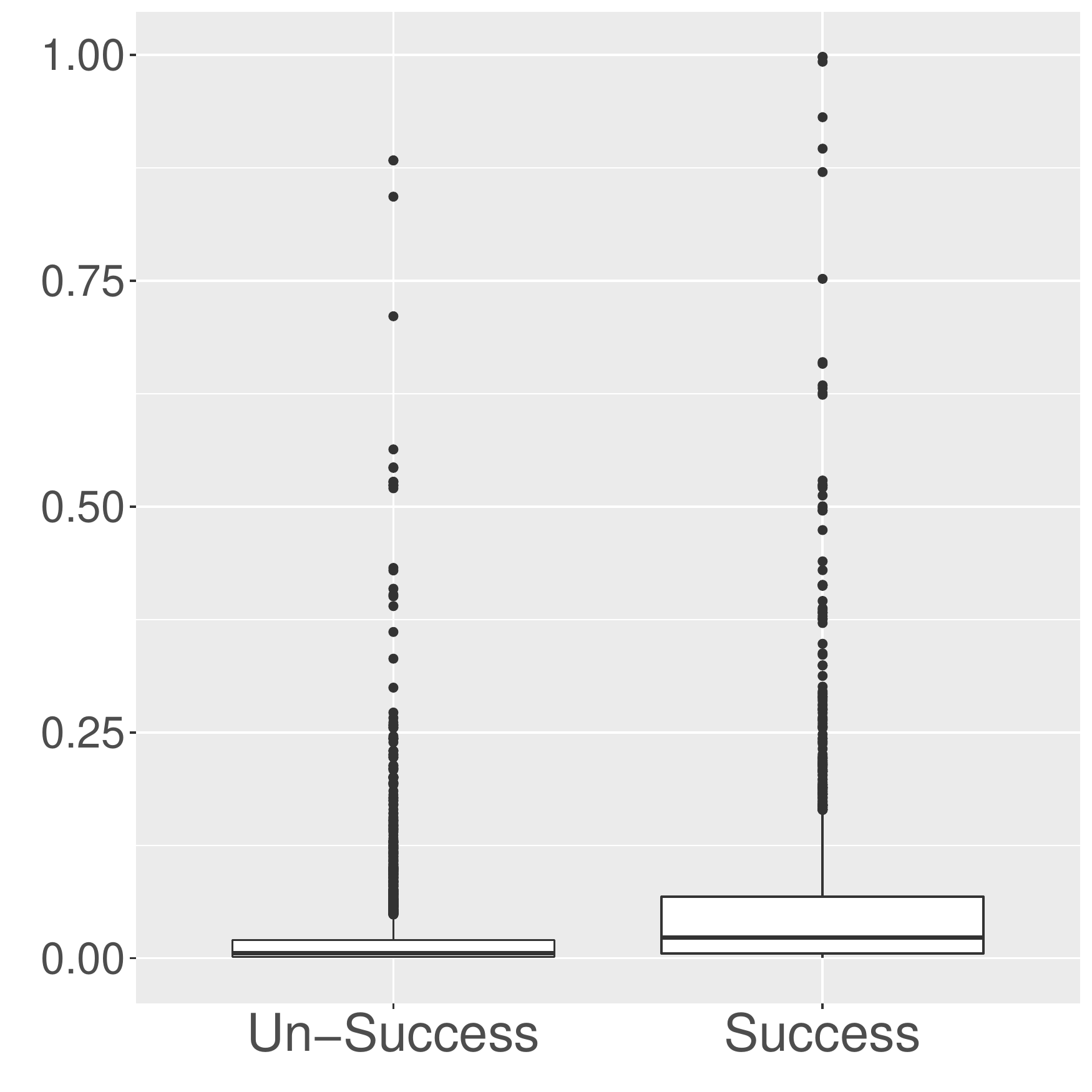} 
	\end{minipage}
}
\vspace*{-.2cm}
\caption{Effect of image complexity measures on successful vs.~unsuccessful {\em decided games} by DM2.}
\label{fig:dm2DSvU}
\end{figure*}

Figures~\ref{fig:dm1SvU} and~\ref{fig:dm2SvU}
show plots of the image complexity  measures in successful vs.~unsuccessful games, for all games 
played by DM1 and DM2. As already noted in Section~\ref{sec:complexity}, 
fewer instances of the target object, fewer objects, and larger
area of the target object correspond to higher chance of the game being 
successful, similarly to what we had noticed for humans.
We observe the same trend when we restrict the analysis to decided games only,  as shown in Figures~\ref{fig:dm1DSvU} and ~\ref{fig:dm2DSvU} for DM1 and DM2, respectively

\begin{figure*}[ht]
\subfloat[no. of instances of target object.]{
	\label{fig:dm1DvUinstance}
	\begin{minipage}{4.7cm}
	\includegraphics[width=4.7cm]{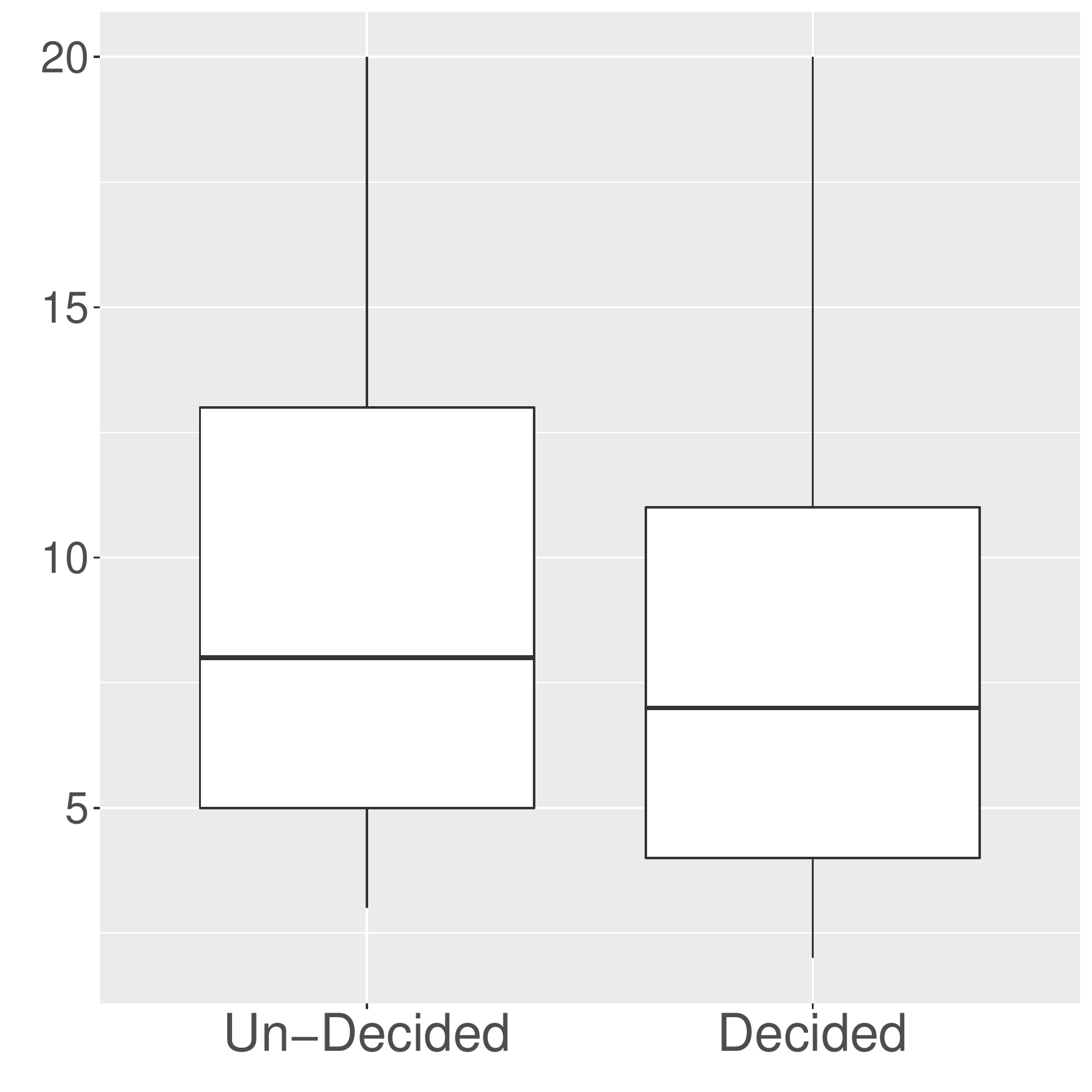} 
	\end{minipage}
}
\subfloat[no. of objects.]{
	\label{fig:dm1DvUobject}
	\begin{minipage}{4.7cm}
	\includegraphics[width=4.7cm]{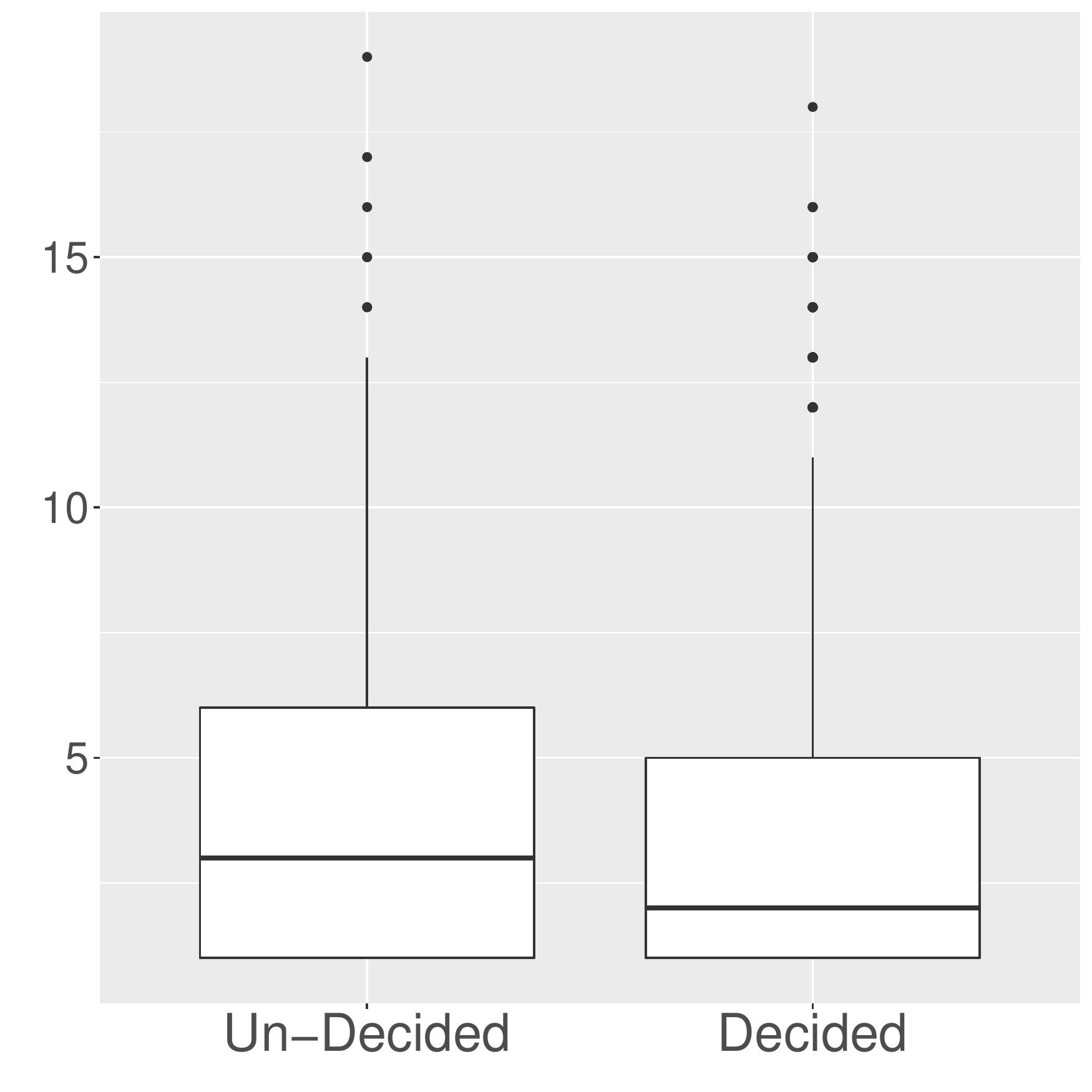} 
	\end{minipage}
}
\subfloat[\% area covered by target object.]{
	\label{fig:dm1DvUarea}
	\begin{minipage}{4.7cm}
	\includegraphics[width=4.7cm]{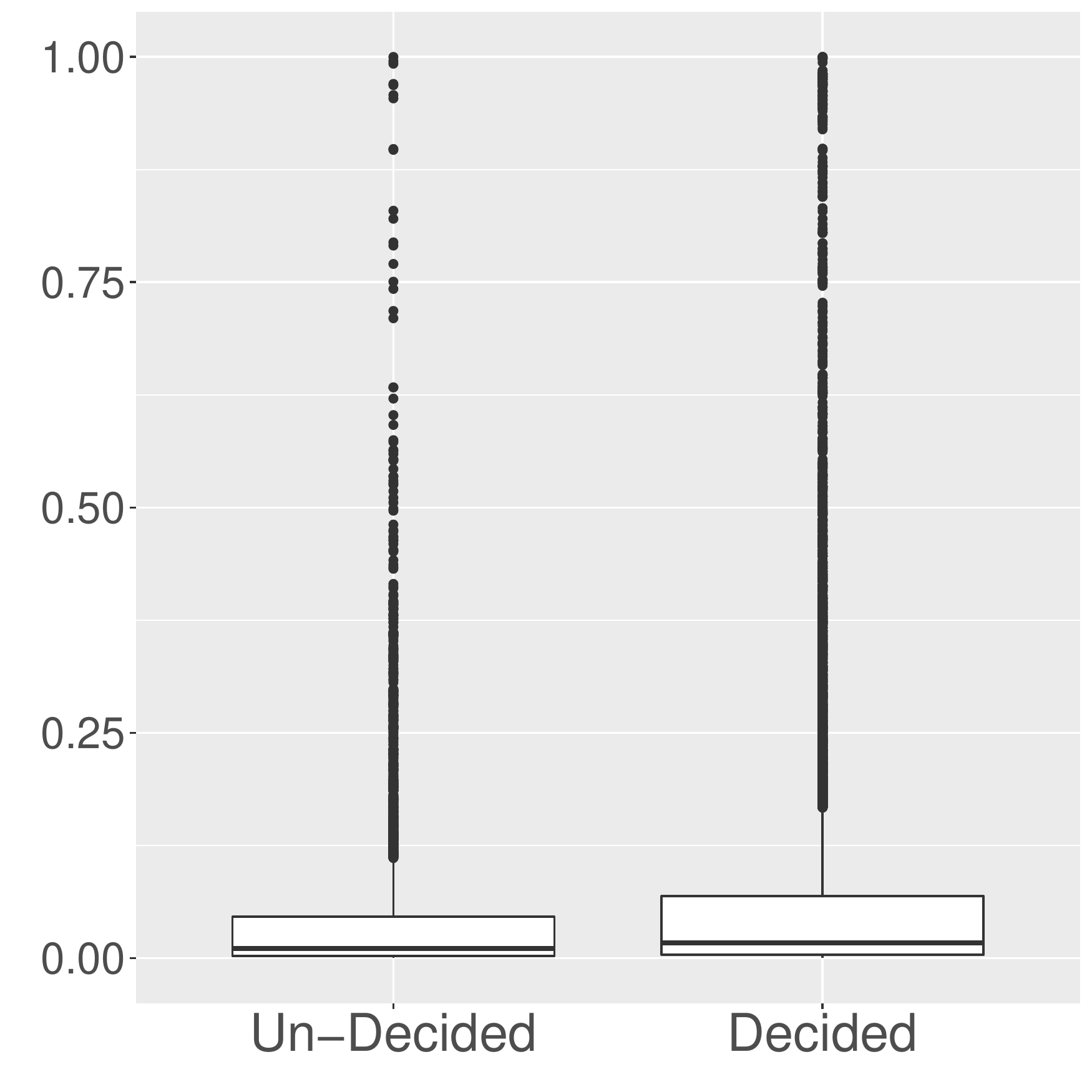} 
	\end{minipage}
}
\vspace*{-.2cm}
\caption{Effect of image complexity measures on decided vs.~undecided games
  played by DM1.}
\label{fig:dm1DvU}
\vspace*{-.8cm}
\end{figure*}

\begin{figure*}[ht]
\subfloat[no. of instances of target object.]{
	\label{fig:dm2DvUinstance}
	\begin{minipage}{4.7cm}
	\includegraphics[width=4.7cm]{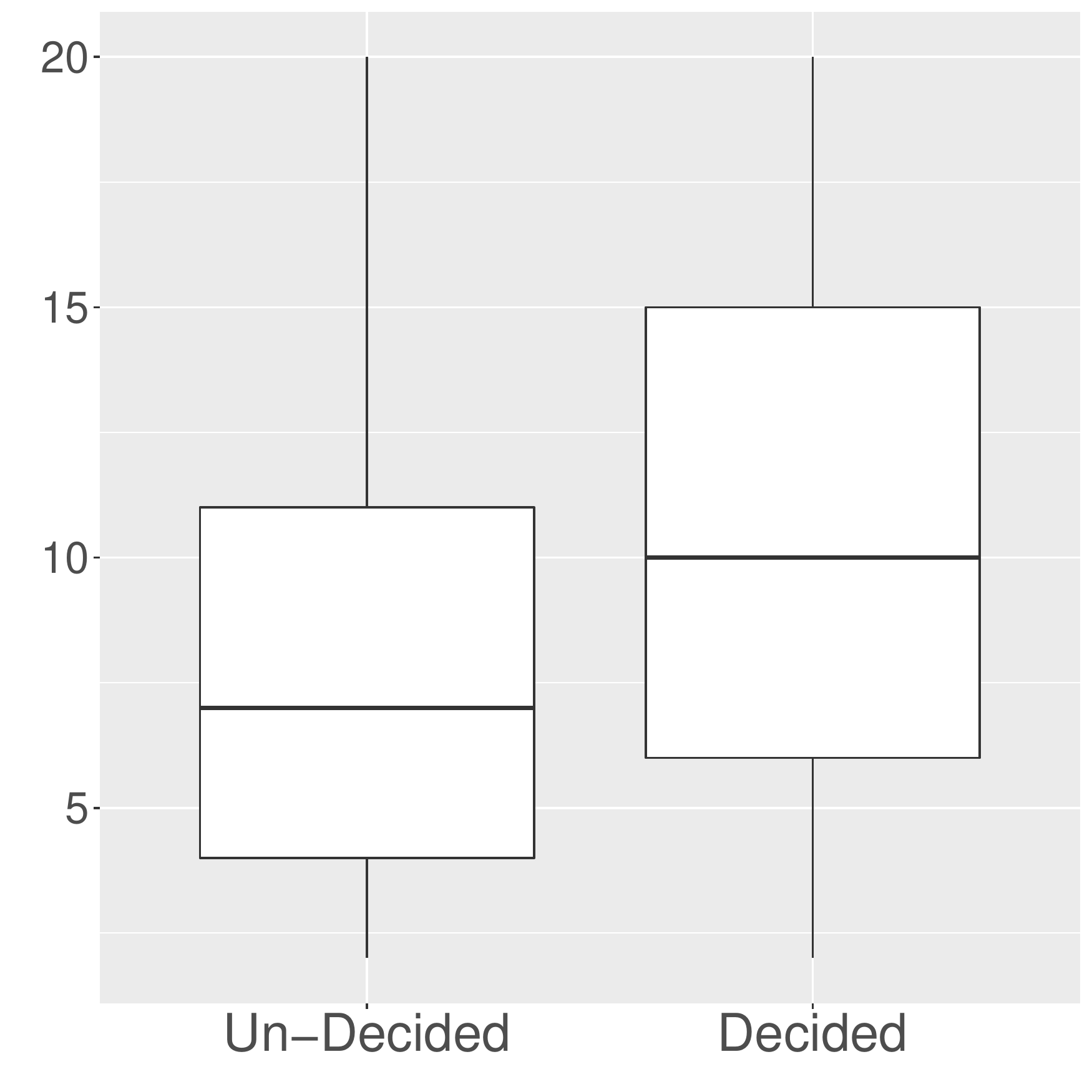} 
	\end{minipage}
}
\subfloat[no. of objects.]{
	\label{fig:dm2DvUobject}
	\begin{minipage}{4.7cm}
	\includegraphics[width=4.7cm]{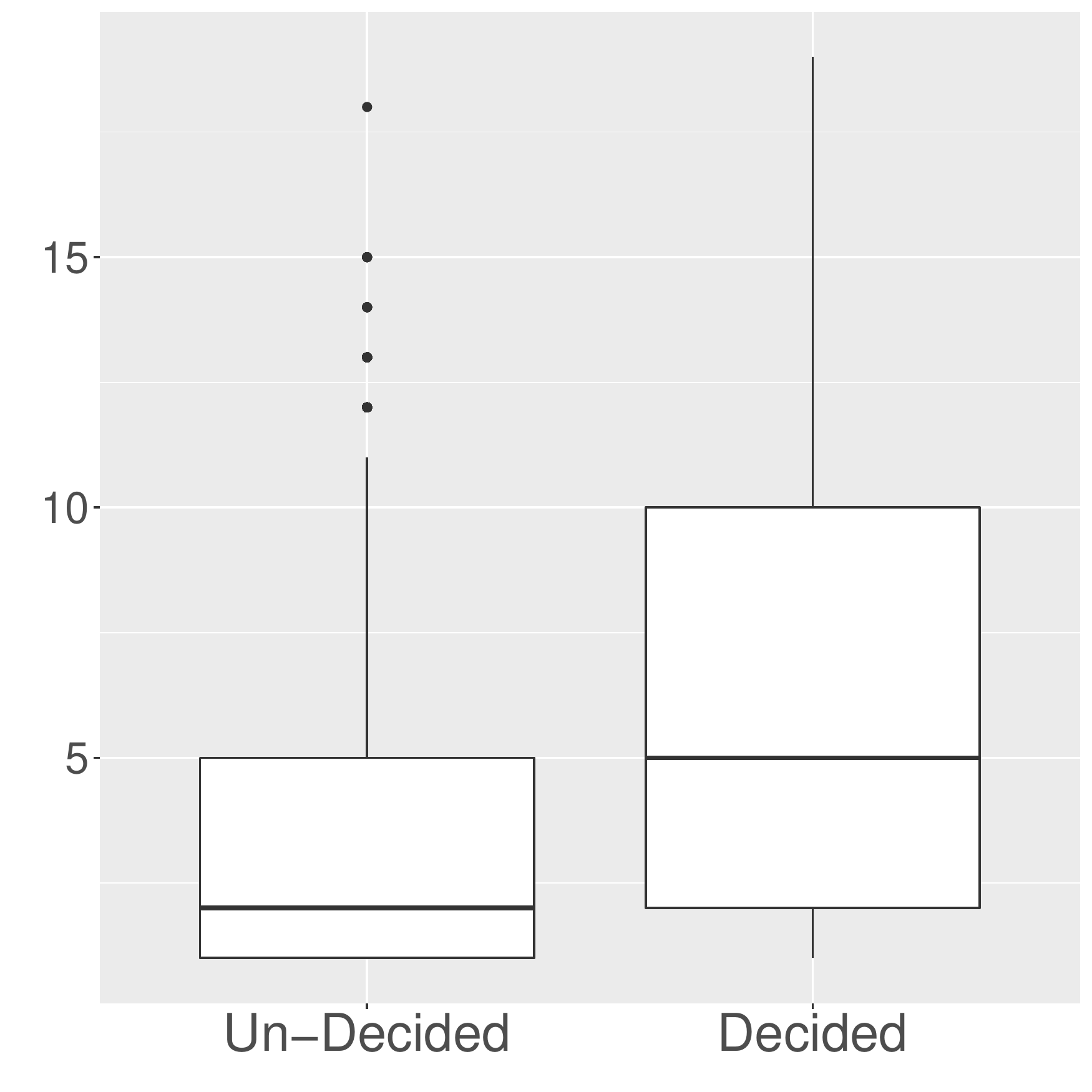} 
	\end{minipage}
}
\subfloat[\% area covered by target object.]{
	\label{fig:dm2DvUarea}
	\begin{minipage}{4.7cm}
	\includegraphics[width=4.7cm]{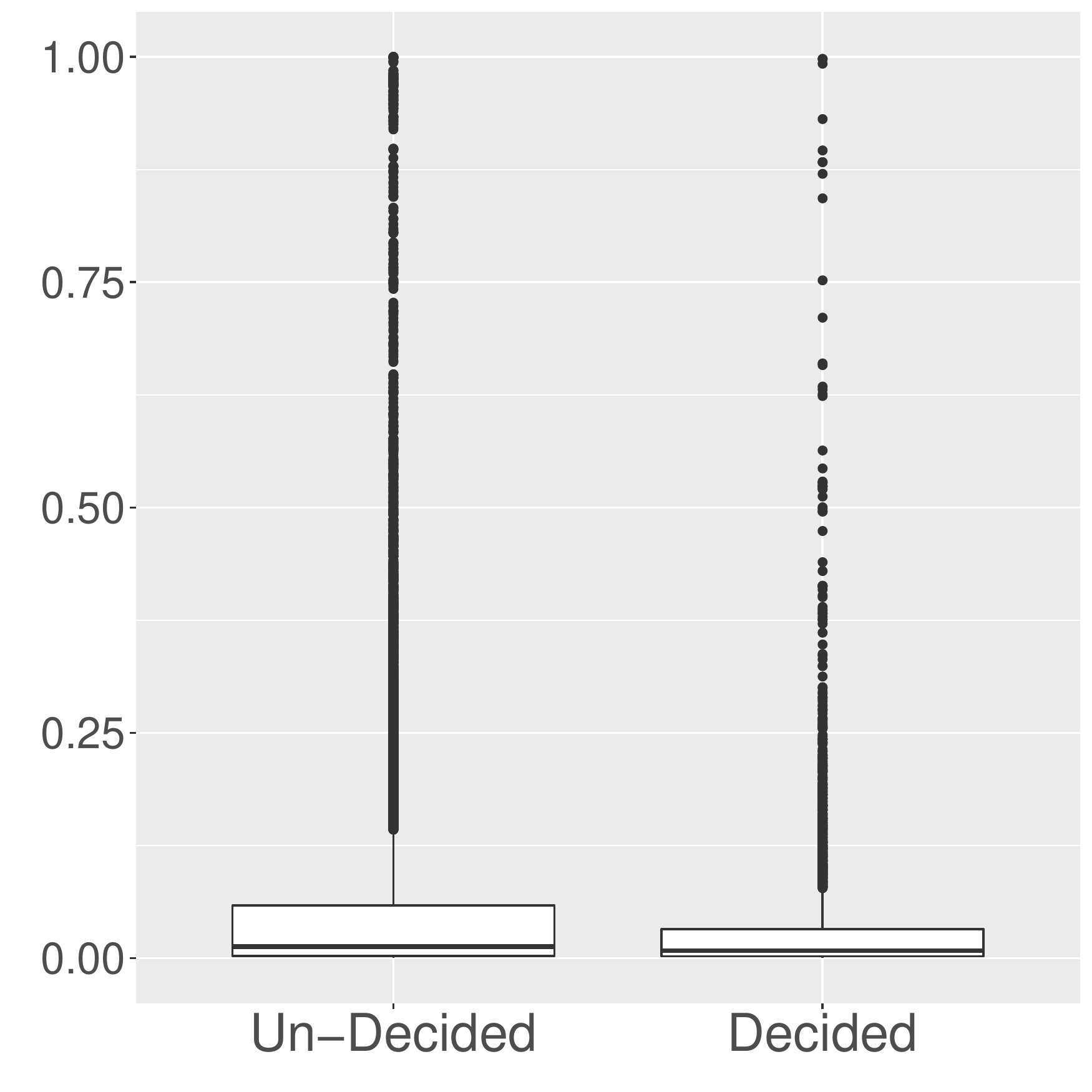} 
	\end{minipage}
}
\vspace*{-.2cm}
\caption{Effect of  image complexity measures on decided vs.~undecided
  games played by DM2.}
\label{fig:dm2DvU}
\end{figure*}

Figures~\ref{fig:dm1DvU} and ~\ref{fig:dm2DvU} compare decided vs.~undecided games played by the two DMs. In this case, we observe a difference: DM1 seems to exploit the image complexity measures in a way similar to humans, as noted earlier. DM2, however, decides more often when there are more instances of the  target object and when the number of objects in the image is higher. Why this is the case remains unclear.

\clearpage


\vspace*{1cm}

\section*{Appendix C: Repeated Questions}

To automatically find the repeated questions in a generated game (see Section~\ref{sec:quality}), we
have used full string matching, i.e., a question is 
considered a repetition only when there is another question in the same
game having exactly the same words. We have considered repetitions for three types of
questions, namely, questions about Object type, Attributes, and Spatial location. Keyword matching is used to decide the type of question that is repetition. For instance, a repeated question is of Object type if it contains keywords such as `dog', `cat', etc. These keywords are
 created using MS-COCO object categories and super-categories, plus the following manually curated list: 
[man, woman, girl, boy, table, meter, bear, cell, phone, wine, glass, racket, baseball, glove, hydrant, drier, kite].

\vspace*{1cm}

\section*{Appendix D: Example Dialogues}

We provide some more examples of successful and unsuccessful games. 

\begin{figure}[h]
	\begin{minipage}{5.3cm}
	\includegraphics[width=5.4cm]{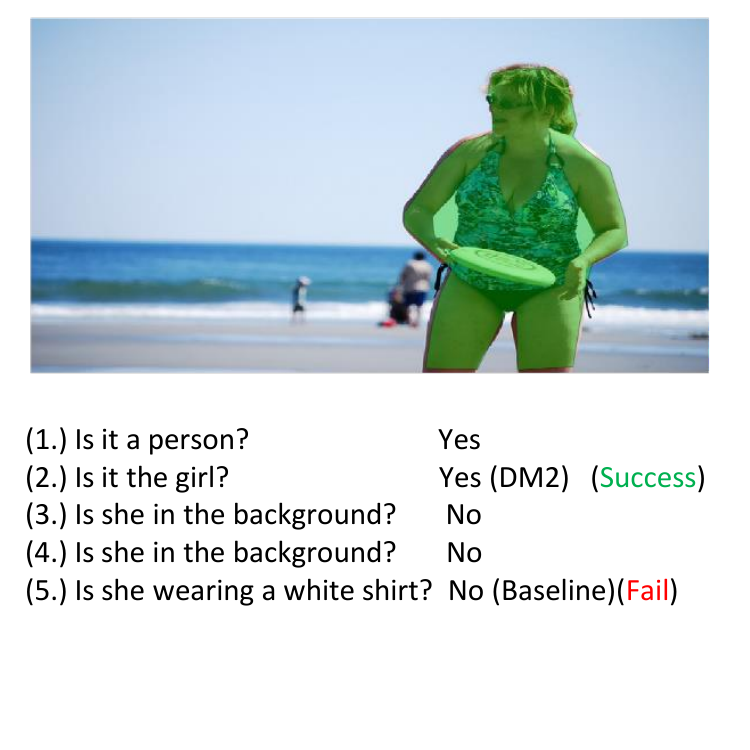}
	\end{minipage}
	\begin{minipage}{5.3cm}
			\includegraphics[width=5.4cm]{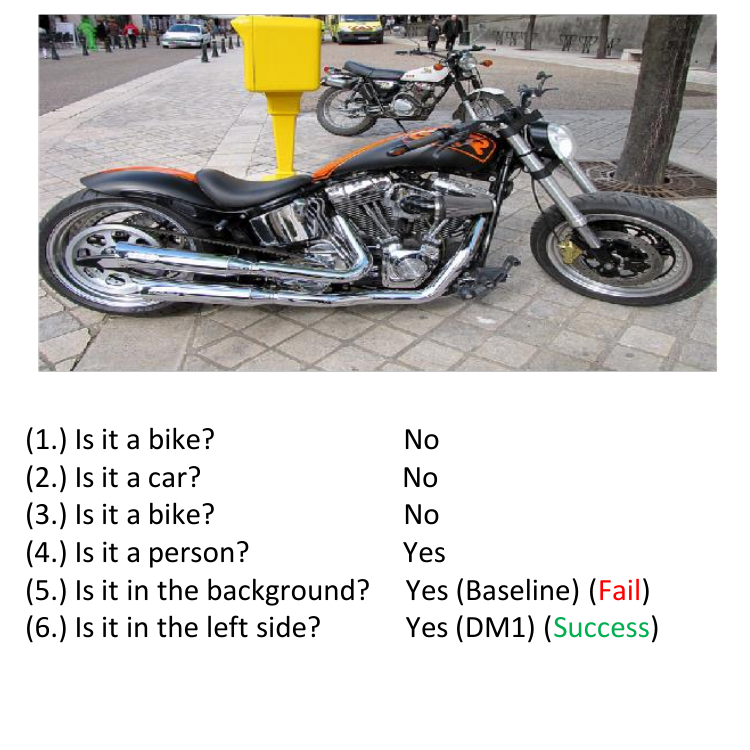} 
	\end{minipage}
	\begin{minipage}{5.3cm}
	\includegraphics[width=5.4cm]{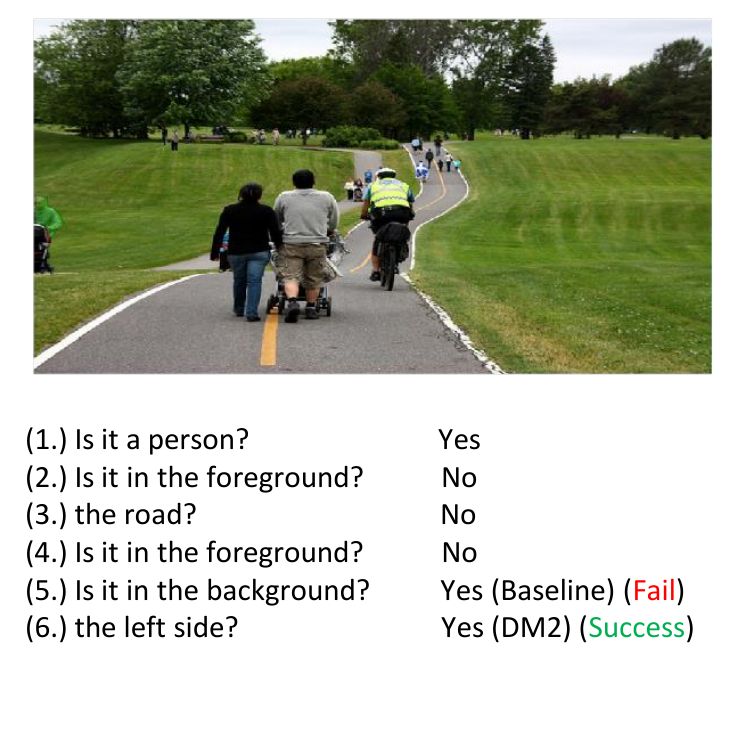} 
	\end{minipage}
\vspace*{-.9cm}	
\caption{Successful games where the DM decides to ask more/fewer questions compared to the Baseline. Some
of the target objects are very small and by allowing to agent to ask extra questions about the spatial location, the Guesser manages to correctly guess the target object.
}
\label{fig:success}
\end{figure}

\vspace*{-.5cm}
\begin{figure}[h]
	\begin{minipage}{5.3cm}
	\includegraphics[width=5.4cm]{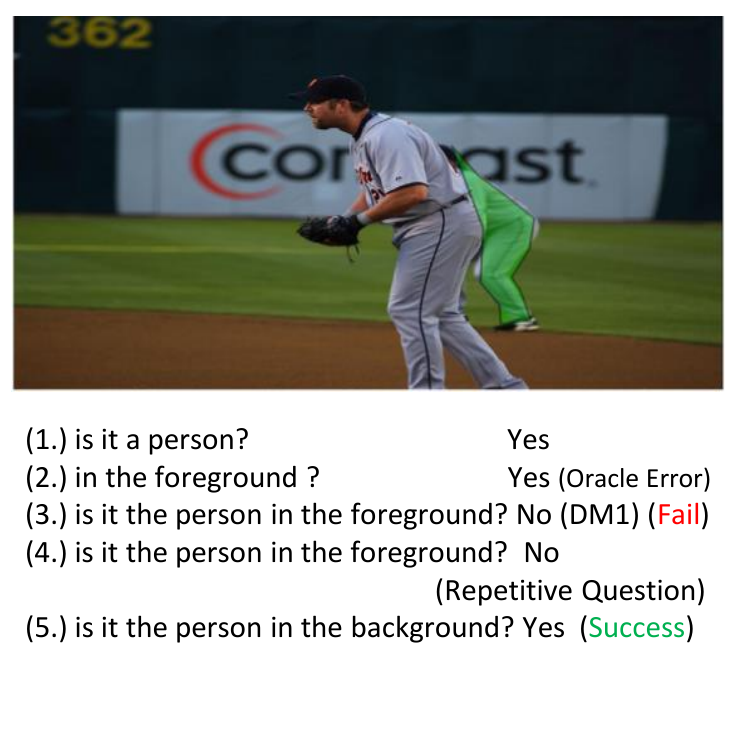} 
	\end{minipage}
	\begin{minipage}{5.3cm}
	\includegraphics[width=5.4cm]{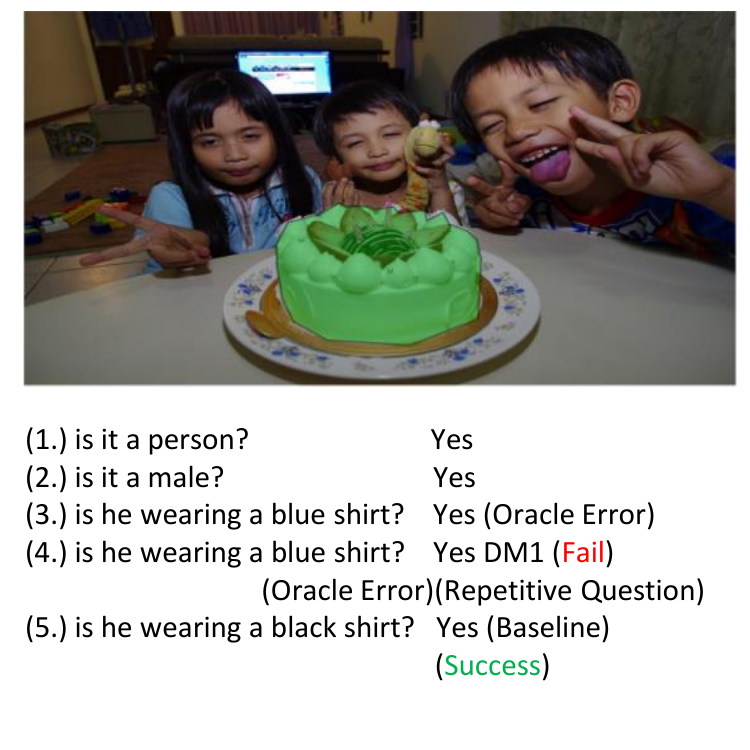} 
	\end{minipage}
	\begin{minipage}{5.3cm}
	\includegraphics[width=5.4cm]{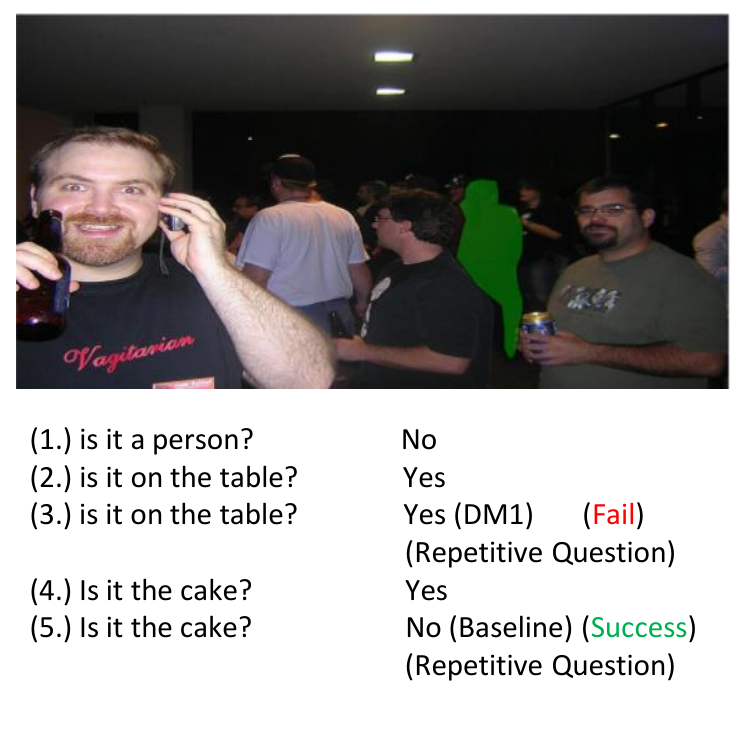} 
	\end{minipage}
\vspace*{-.9cm}	
\caption{Unsuccessful games played by DM1, where the system decides to stop asking earlier than the baseline. These failures seem to occur due to the problems by QGen (which generates repeated questions) or the Oracle (which provides a wrong answer).}
\label{fig:failure1}
\end{figure}

\vspace*{-1cm}
\begin{figure}[h] \centering
	\begin{minipage}{5.3cm}
	\includegraphics[width=5.4cm]{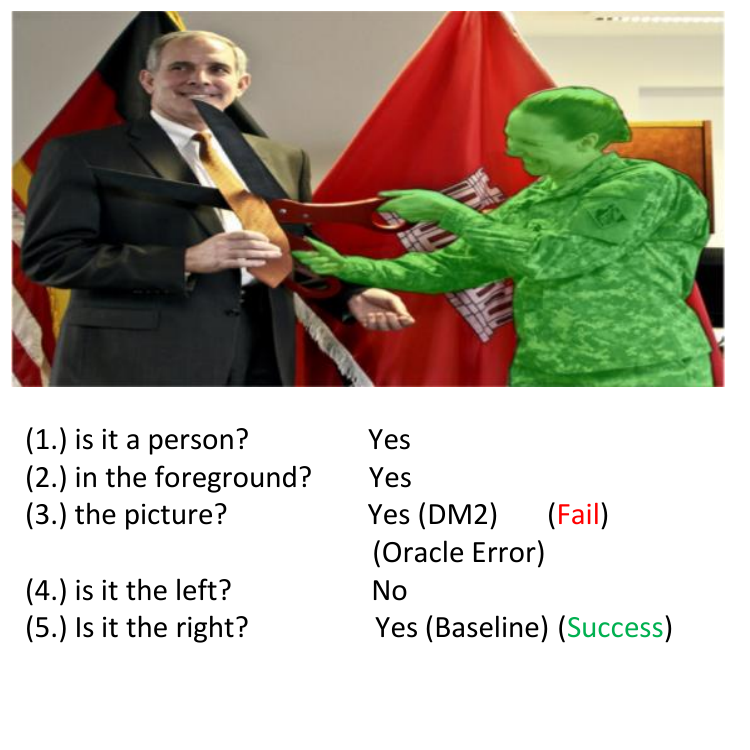} 
	\end{minipage}
	\begin{minipage}{5.3cm}
	\includegraphics[width=5.4cm]{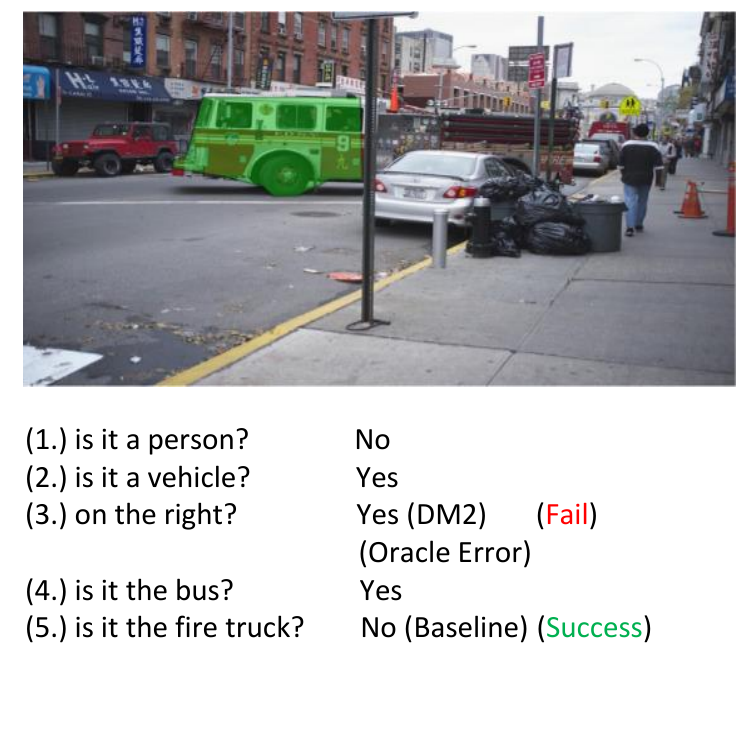} 
	\end{minipage}
\vspace*{-.9cm}	
\caption{Unsuccessful games played by DM2, where the system decides to stop asking earlier than the baseline. DM2 seems to be overconfident. Again, failures are often affected by Oracle errors.
}
\label{fig:failure2}
\end{figure}

\end{document}